\documentclass[10pt]{article}

\usepackage[a4paper,margin=1in]{geometry}
\usepackage{amsmath,amssymb,amsfonts}
\usepackage{graphicx}
\usepackage{booktabs}
\usepackage{caption}
\usepackage{subcaption}
\usepackage{siunitx}
\usepackage{textcomp}
\usepackage{algorithmic}
\usepackage[hidelinks]{hyperref}
\usepackage{cite} % or natbib/biblatex if you prefer
\usepackage{url}
\usepackage{authblk}

\def\BibTeX{{\rm B\kern-.05em{\sc i\kern-.025em b}\kern-.08em
    T\kern-.1667em\lower.7ex\hbox{E}\kern-.125emX}}

\begin{document}
\title{Unsupervised Anomaly Detection of Diseases in the Female Pelvis for Real-Time MR Imaging}
\author[1,2,*]{A.~Knupfer}
\author[3]{J.~P.~M\"uller}
\author[1]{J.~A.~Verdera}
\author[1]{M.~Fenske}
\author[1]{C.~S.~Mathy}
\author[1]{S.~Tripathy}
\author[1,4]{S.~Arndt}
\author[1]{M.~May}
\author[1]{M.~Uder}
\author[5]{M.~W.~Beckmann}
\author[5]{S.~Burghaus}
\author[2]{J.~Hutter}

\affil[1]{Radiological Institute, University Hospital Erlangen (UKER), Erlangen, Germany}
\affil[2]{Institute of Information Processing, Leibniz University Hannover, Hannover, Germany}
\affil[3]{Image Data Exploration and Analysis Lab, Friedrich-Alexander University Erlangen–N\"urnberg (FAU), Erlangen, Germany}
\affil[4]{Medical Center for Information and Communication Technology, UKER}
\affil[5]{Institute of Women's Health, UKER}

\affil[*]{Corresponding author: anika.knupfer@fau.de}
\date{}
\maketitle

\begin{abstract}
Pelvic diseases in women of reproductive age represent a major global health burden, with diagnosis frequently delayed due to high anatomical variability, complicating MRI interpretation. Existing AI approaches are largely disease-specific and lack real-time compatibility, limiting generalizability and clinical integration. To address these challenges, we establish a benchmark framework for disease- and parameter-agnostic, real-time–compatible framework for unsupervised anomaly detection in pelvic MRI. The method designs a residual variational autoencoder trained exclusively on healthy sagittal T2-weighted scans acquired across diverse imaging protocols to model normal pelvic anatomy. During inference, reconstruction error heatmaps highlight deviations from learned healthy structure, enabling detection of pathological regions without labeled abnormal data. The model is trained on $294$ healthy scans and augmented with diffusion-generated synthetic data to improve robustness. Quantitative evaluation on the publicly available Uterine Myoma MRI Dataset (UMD) yields an average area under the curve (AUC) value of $0.736$, with $0.828$ sensitivity and $0.692$ specificity. Additional inter-observer clinical evaluation extends analysis to endometrial cancer, endometriosis, and adenomyosis, revealing the influence of anatomical heterogeneity and inter-observer variability on performance interpretation. With a reconstruction time of $\approx92.6$ frames per second, the proposed framework establishes a baseline for unsupervised anomaly detection in the female pelvis and towards future integration into real-time MRI. Code is available upon request (https://github.com/AniKnu/UADPelvis), prospective data sets are available for academic collaboration.

\end{abstract}

\section{Introduction}
\label{sec:introduction}
\begin{figure}[!t]
\centerline{\includegraphics[width=0.63\columnwidth]{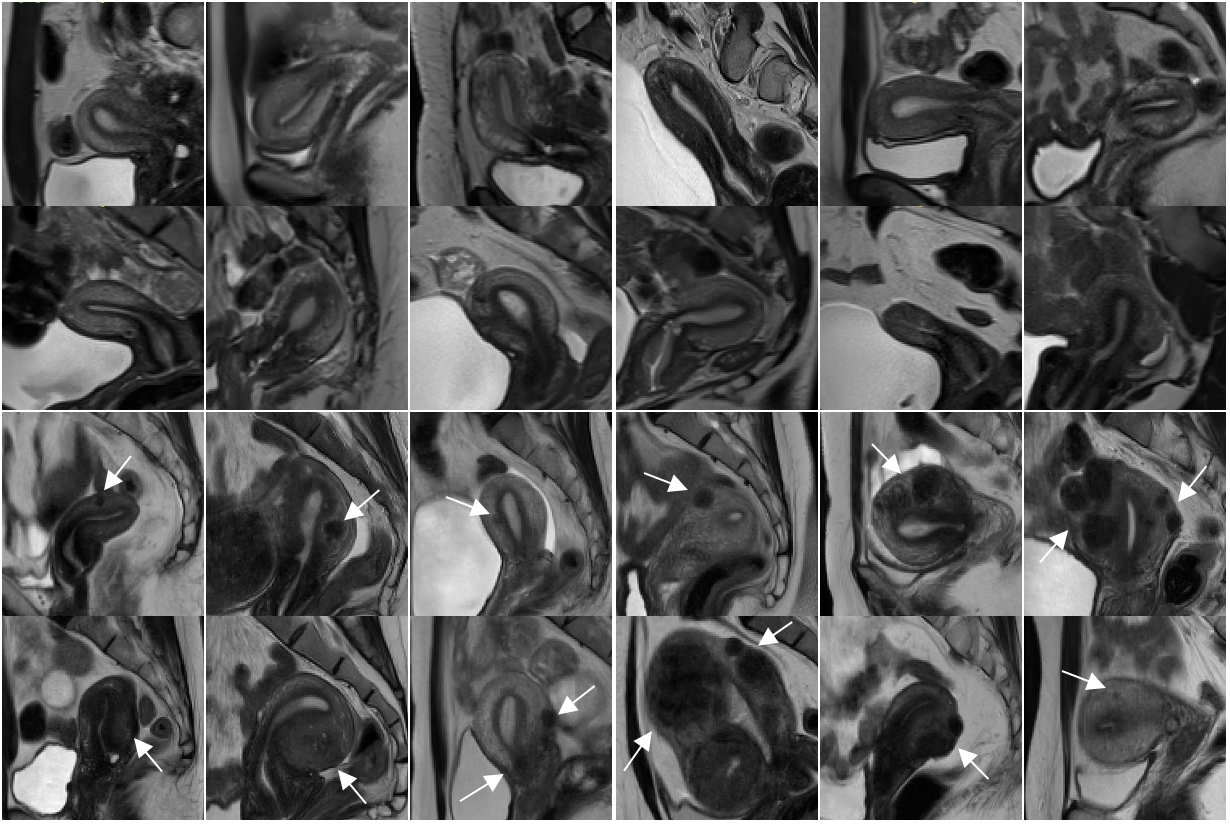}}
\caption{The upper rows depict healthy uteri, while the lower rows represent pathological cases, with arrows indicating anomalies such as myomas, cysts, and endometriosis, highlighting the anatomical variability of the female pelvis.}
\label{HealthyUnhealthy}
\end{figure}

Pelvic diseases pose a significant health challenge for women of reproductive age, affecting approximately $40~\%$ of this population~\cite{gao2025rising} and contributing to $4.5~\%$ of the global disease burden~\cite{Sebastianojnumed.124.267546}. These conditions include uterine myomas, adenomyosis, endometriosis, ovarian tumors, and endometrial or cervical cancer, among others — each presenting unique diagnostic challenges due to their variable appearance in medical imaging~\cite{sudderuddin2014mri}. This variability often results in diagnostic delays, misclassification, unnecessary suffering, and, in some cases, avoidable surgical interventions. For example, endometriosis affects about $10~\%$ of women, yet is often misdiagnosed, with an average delay of eight years and nearly $75~\%$ of patients initially misdiagnosed~\cite{hudelist2012diagnostic, bulun2019}. An important factor contributing to the complexity of pelvic disease diagnosis is the substantial anatomical variability of the uterus. While the anteverted and anteflexed positions are most common in healthy uteri, other orientations, such as retroverted, retroflexed, or intermediate, represent normal variants~\cite{sudderuddin2014mri} (see Fig.~\ref{HealthyUnhealthy}).

Currently, detecting and localizing pelvic diseases is time-consuming and costly. While ultrasound is widely used for detection, its interpretability can vary and acoustic shadowing potentially affecting lesion size estimation or risking misinterpretations~\cite{cai2024artificial}. In contrast, MRI plays a critical role in more complex cases due to its superior spatial resolution, better visualization of deep pelvic structures, improved soft tissue contrast and characterization, broader field of view, and reduced operator dependence, ultimately enabling more accurate diagnosis and treatment planning~\cite{mahmoud2025comparative, tong2024best}. However, the lack of standardized protocols and often oblique position of the uterus resulting in unusual planes for pelvic MRIs complicates matters further, as diagnoses depend on the clinician's expertise and on the images acquired at the time of the scan~\cite{nougaret2022mri}. A real-time MRI scan that can quickly detect and localize pelvic diseases could significantly improve diagnostic speed and accuracy, reduce risks, and improve patient outcomes.

The integration of Artificial Intelligence (AI) into relatively invariant areas such as brain or chest imaging offers the potential to address these challenges by assisting radiologists in identifying potential anomalies, improving diagnostic accuracy, and supporting clinical decision-making~\cite{satushe2025ai}. However, these methods often fail when applied to anatomically variable regions like the abdomen and pelvis, where high inter-patient variability poses unique challenges for generalization, emphasizing the need for specialized approaches tailored to these complex areas~\cite{cai2025artificial}. This integration could, in the case of female reproductive health, improve treatment outcomes, and a more personalized approach~\cite{aftab2025artificial, dungate2024assessing}. However, the considerable variability of both normal and abnormal tissues in the female pelvis described above also presents a significant challenge for current machine learning models~\cite{zabihollahy2021fully}. Given the wide range of pelvic disease manifestations, it is not feasible to capture all possible abnormalities in a data set for supervised learning, which has been contributed to date in isolated AI research with specific disease foci~\cite{dungate2024assessing}. Furthermore, inter-observer variability in the interpretation of pelvic MRI images can lead to significant differences in expert annotations, reflecting clinical reality and complicating the creation of consistent reference standards for supervised deep learning~\cite{karimi2020deep}. To address these limitations, unsupervised anomaly detection (UAD) has emerged as a promising approach in medical imaging. UAD allows AI models to learn the distribution of normal data, essentially, healthy anatomy, without the need to label abnormal data. This approach mirrors human learning, wherein anomalies are recognized by first understanding what is "normal" and then identifying deviations. Training deep learning models with healthy data enables the resulting models to detect anomalies by highlighting areas where the model's reconstruction deviates from the inputs~\cite{baur2021autoencoders, baur2021modeling,raad2023unsupervised,baugh2023many,behrendt2024patched,patsanis2023comparison}.

\textbf{Contributions.} This study presents a baseline for disease-independent deep learning models for UAD in the female pelvis. To account for the considerable diversity of acquisition parameters, the model is not subject to protocol restrictions but is trained on a variety of T2-weighted (T2w) sagittal pelvic examinations acquired across field strengths. Beyond quantitative performance evaluation, this work demonstrates how anatomical diversity and expert interpretation influence anomaly detection results, providing important context for evaluating UAD methods in pelvic MRI. The demonstrated low latency inference approach paves the way for future integration into real-time MRI workflows. 

\section{Related Work}
Recent studies have applied deep learning to anomaly detection in the female pelvis across imaging modalities. In ultrasound, Yang et al.~\cite{yang2023real} proposed a real-time supervised approach for uterine myoma detection, while Huo et al.~\cite{huo2023artificial} introduced a two-stage Convolutional Neural Network (CNN) to support less experienced technicians. Shahzad et al.~\cite{shahzad2023automated} developed a dual-path CNN for myoma classification, and Chen et al.~\cite{chen2020deep} used CNNs to assess myometrial invasion in endometrial cancer. In MRI, Hodneland et al.~\cite{hodneland2021automated} segmented endometrial tumors using a CNN, extracting texture features for radiomic profiling, though with reduced performance on non-endometrioid tumors. In~\cite{figueredo2024automatic}, Figueredo et al. proposed a five-step deep learning pipeline for detecting rectosigmoid endometriosis in T2w MRI, achieving $96.67~\%$ patient classification and a Dice score of $65.44~\%$. However, both mentioned MRI studies were limited by small data sets and dependency on specific imaging protocols and disease characteristics. 

UAD specifically has emerged as a promising alternative in areas such as in brain MRI where Baur et al.~\cite{baur2021modeling} demonstrated that convolutional autoencoders can outperform conventional methods in lesion detection. Raad et al.~\cite{raad2023unsupervised} achieved $83~\%$ accuracy in detecting neonatal encephalopathy using an unsupervised deep autoencoder. While these approaches highlight the potential of UAD, their application to pelvic imaging remains largely unexplored.

Generative models, including variational autoencoders (VAE), generative adverserial networks (GAN), and diffusion models, have already been extensively investigated for UAD. Previous work has demonstrated their potential for various organs, including GAN-based approaches for prostate MRI~\cite{patsanis2023comparison}, autoencoder benchmarks for detecting multiple sclerosis lesions highlighting the advantages of VAEs~\cite{baur2021autoencoders}, and diffusion-based reconstruction of local brain anatomy~\cite{behrendt2024patched}. While image-conditioned diffusion models can generate realistic pseudo-healthy reconstructions~\cite{baugh2024image}, self-supervised anomaly generation methods face an inherent tradeoff between realism and detectability due to challenges in plausibly integrating synthetic lesions~\cite{baugh2023many}. Self-supervised approaches often rely on synthetic anomaly generation through cut-and-paste strategies or structured noise, but face an inherent tradeoff between realism and detectability of the generated lesions~\cite{baugh2023many,p2023confidence}. Despite these advances, UAD remains largely unexplored in female pelvic MRI, where high anatomical variability and non-standardized acquisition protocols pose substantial challenges for robust and generalizable model development.

\section{Materials and Methods}
\label{sec:MatMethods}
\begin{figure*}[!t]
\centerline{\includegraphics[width=0.95\textwidth]{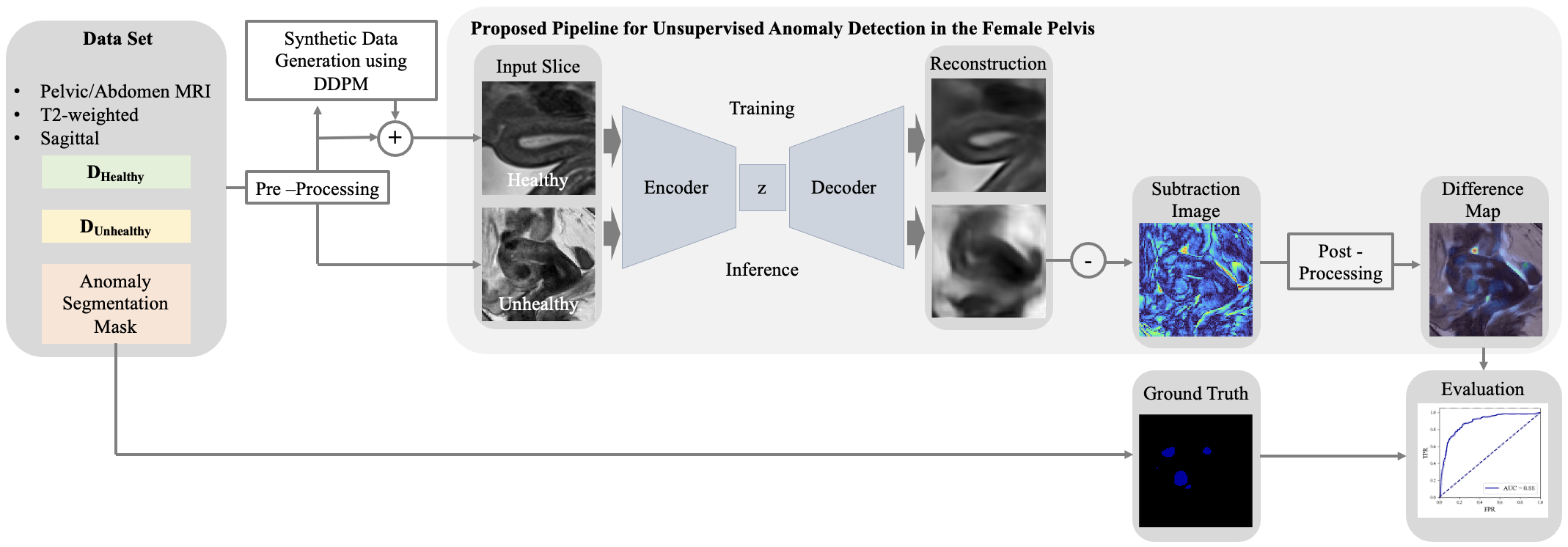}}
\caption{Overview of the proposed unsupervised pelvic MRI anomaly detection pipeline. Preprocessed sagittal T2w scans are augmented using an in-house DDPM and used to train a residual variational autoencoder on healthy anatomy. During inference, reconstruction errors are post-processed and overlaid to highlight anomalies. Evaluation is based on ground-truth segmentations.}
\label{pipeline}
\end{figure*}
The following methodology establishes a baseline approach for UAD in pelvic MRI, prioritizing generalizability across acquisition protocols and disease types. The full UAD pipeline is depicted in Fig.~\ref{pipeline}, encompassing pre-processing, synthetic data generation, training of the variational autoencoder (VAE), post-processing, and performance evaluation.

\subsection{Data Sets}
To construct the training data set D\textsubscript{\textit{Healthy}}, a PACS query was performed to identify pelvic MRI examinations acquired in female patients from the local university hospital that included a sagittal T2-weighted view of the uterus, as this is the most commonly acquired contrast and orientation. This query spanned a wide range of clinical indications, including but not limited to suspected endometriosis, uterine cancer, rectal pathology, and hip evaluation. A total of $800$ recent examinations fulfilling these criteria were identified, retrospectively reviewed by an experienced radiologist and categorized into four groups: no uterus, healthy uterus, minor anomalies, and major anomalies. Only cases with normal uterine anatomy were included, resulting in $294$ healthy T2-weighted scans. All data were fully anonymized in line  with the institutional ethics guidelines (ethics number: 24-304-Br). The data set exhibits substantial acquisition and physiological variability due to the absence of standardized pelvic MRI protocols. Variations include field strengths ($0.55$~–~$3~\mathrm{T}$), sequence types (TSE, HASTE), matrix sizes ($208 \times 208$~–~$832 \times 832$), scanner vendors (Siemens, Philips), contrast agent usage, menstrual cycle phase, bladder filling, and patient age (mean $30.01$ years). This diversity improves the generalizability of the model and supports realistic synthetic data generation. 

The evaluation data set D\textsubscript{\textit{Unhealthy}} comprises $242$ pathological sagittal T2w pelvic MRI scans from two sources. The public Uterine Myoma MRI Dataset (UMD)~\cite{pan2024large, pan2023} provides standardized acquisitions obtained on Philips scanners with fixed protocols and motion-reduction techniques, enabling reproducible quantitative evaluation. UMD annotations were generated by $11$ experts using ITK-SNAP~\cite{py06nimg}, labeling uterine structures, myomas, and Nabothian cysts according to FIGO classification. Additionally, $33$ in-house pathological cases were selected based on validated radiological reports and ICD codes, including uterine myomas, endometrial cancer, endometriosis, and adenomyosis. Lesion ground truth (GT) segmentations were independently generated by two experienced radiologists and one unexperienced observer to assess annotation-related variability. D\textsubscript{\textit{Unhealthy}} includes patients with a mean age of $48.80$ years. The two subsets are referred to as D\textsubscript{\textit{Unhealthy(UMD)}} and D\textsubscript{\textit{Unhealthy(in-house)}}, and collectively as D\textsubscript{\textit{Unhealthy}}.

\subsection{Pre-Processing}
\begin{figure}[!t]
\centerline{\includegraphics[width=0.65\columnwidth]{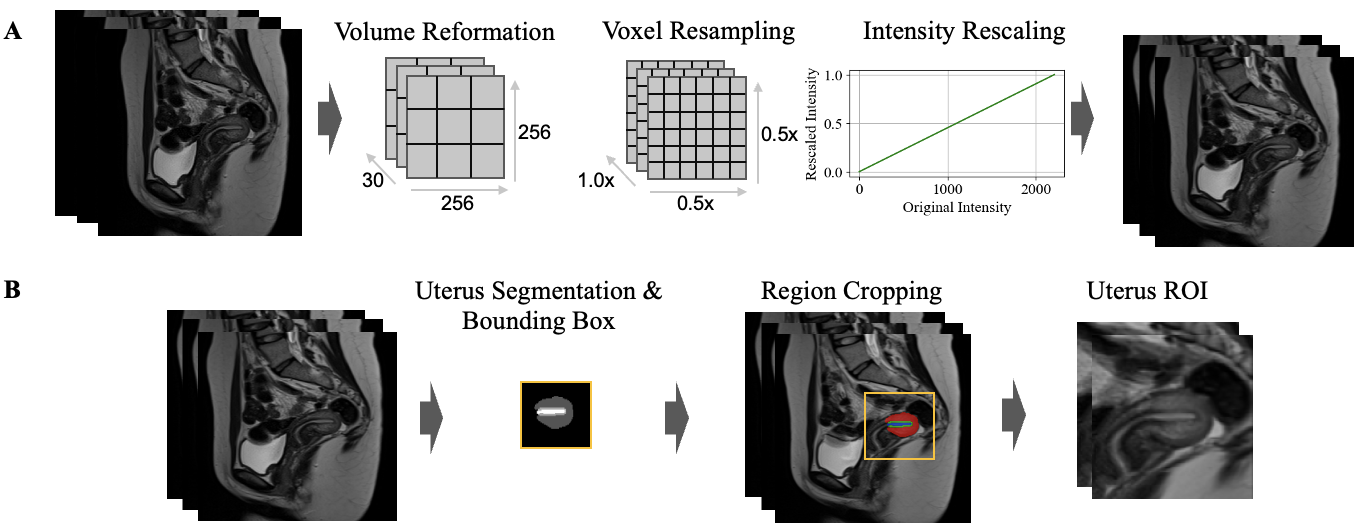}}
\caption{Pre-processing pipeline for standardized model input, including voxel and intensity normalization, anatomical segmentation using an attention-based 3D U-Net~\cite{tripathydeep}, and segmentation-driven bounding box cropping.}
\label{preprocess}
\end{figure}
MRI volumes from both D\textsubscript{\textit{Healthy}} and D\textsubscript{\textit{Unhealthy}} were converted from DICOM format~\cite{dicomstandard} to NIfTI~\cite{nifti} using SimpleITK~\cite{yaniv2018simpleitk}. Volumes were resampled to a uniform voxel spacing of $0.5~\mathrm{mm}\times~0.5~\mathrm{mm}\times~1~\mathrm{mm}$, resized to $256~\times~256~\times~30$ voxels, and intensity-normalized to $[0,1]$, ensuring spatial and numerical consistency. To focus on the uterine region, volumes were cropped using segmentation-guided bounding boxes (Fig.~\ref{preprocess}). An in-house, deeply supervised attention-based 3D U-Net~\cite{tripathydeep}, trained on $80$ healthy uterine MRI volumes across multiple field strengths ($0.55$~–~$3~\mathrm{T}$), segmented the endometrium, junctional zone (JZ), and myometrium in D\textsubscript{\textit{Healthy}} and D\textsubscript{\textit{Unhealthy(in-house)}}. Largest connected component filtering removed spurious regions, and segmentations were manually refined with ITK-SNAP~\cite{py06nimg}. For the UMD data set, provided uterus masks were used directly, with only voxel resampling applied. Bounding boxes of size $96~\times~96~\times~n$ (with $n$ denoting slices containing uterine tissue) were generated, enclosing the uterus. MRI volumes, segmentation masks, and anomaly annotations were cropped accordingly and used as input for anomaly detection and data generation. Following exclusion of $91$ UMD cases due to incomplete uterine coverage, D\textsubscript{\textit{Unhealthy}} comprised $242$ volumes ($2{,}966$ UMD, $401$ in-house slices) and D\textsubscript{\textit{Healthy}} comprised $294$ volumes ($2{,}820$ slices).

\subsection{Diffusion Model for Synthetic Image Generation}
\label{sec:DiffModels}
To address data scarcity and variability in healthy female anatomy, a DDPM is employed to generate synthetic T2w pelvic MR images~\cite{ho2020denoising}. The model architecture is based on the Hugging Face UNet2DModel following the U-Net design proposed by Ronneberger et al.~\cite{ronneberger2015u, von-platen-etal-2022-diffusers} and integrated into a DDPM framework. Healthy 3D volumes are decomposed into 2D slices, normalized, and used to train an unconditioned diffusion model according to established implementations~\cite{muller2025diffusing}. Synthetic images are generated from a pre-trained checkpoint using varying random seeds for diversity.

To mitigate risks of model memorization, Structural Similarity Index (SSIM) is employed to quantify the perceptual similarity between real and synthetic image pairs. Synthetic images are compared to their nearest real counterpart, and samples with $\mathrm{SSIM} > 0.35$ are excluded. This empirically chosen threshold ensures sufficient visual dissimilarity while preserving anatomical realism. Applying this criterion resulted in a final set of $1{,}400$ synthetic pelvic MRI slices. Qualitative evaluation is conducted by inspecting randomly sampled synthetic images to assess anatomical plausibility, variability in uterine orientation, contrast characteristics, and the presence of realistic acquisition artifacts. The retained synthetic images are combined with the original healthy data to form the expanded training set D\textsubscript{\textit{Healthy\_Gen}}.

\subsection{Unsupervised Anomaly Detection Pipeline}
\subsubsection{Data Loader} To prevent data leakage, scans from the same patient were not split across training and validation. An $80{:}20$ train/validation split was applied to D\textsubscript{\textit{Healthy}}, ensuring anatomical variability in the validation data, while synthetic samples were used exclusively for training. No data from D\textsubscript{\textit{Healthy}} is used for inference to preserve the integrity of the limited training data. Volumes were sliced along the $z$-axis into $96 \times 96$ grayscale 2D images. Each training slice was augmented up to three times using horizontal flips ($p=0.9$), vertical flips ($p=0.7$), and Contrast Limited Adaptive Histogram Equalization (CLAHE, $p=0.7$, clip limit $0.03$)~\cite{reza2004realization}. All slices were shuffled and a batch size of $32$ was used to account for the heterogeneity of the data set.

\subsubsection{Model Architecture and Training} 
\begin{figure}[!t]
\centerline{\includegraphics[width=\columnwidth]{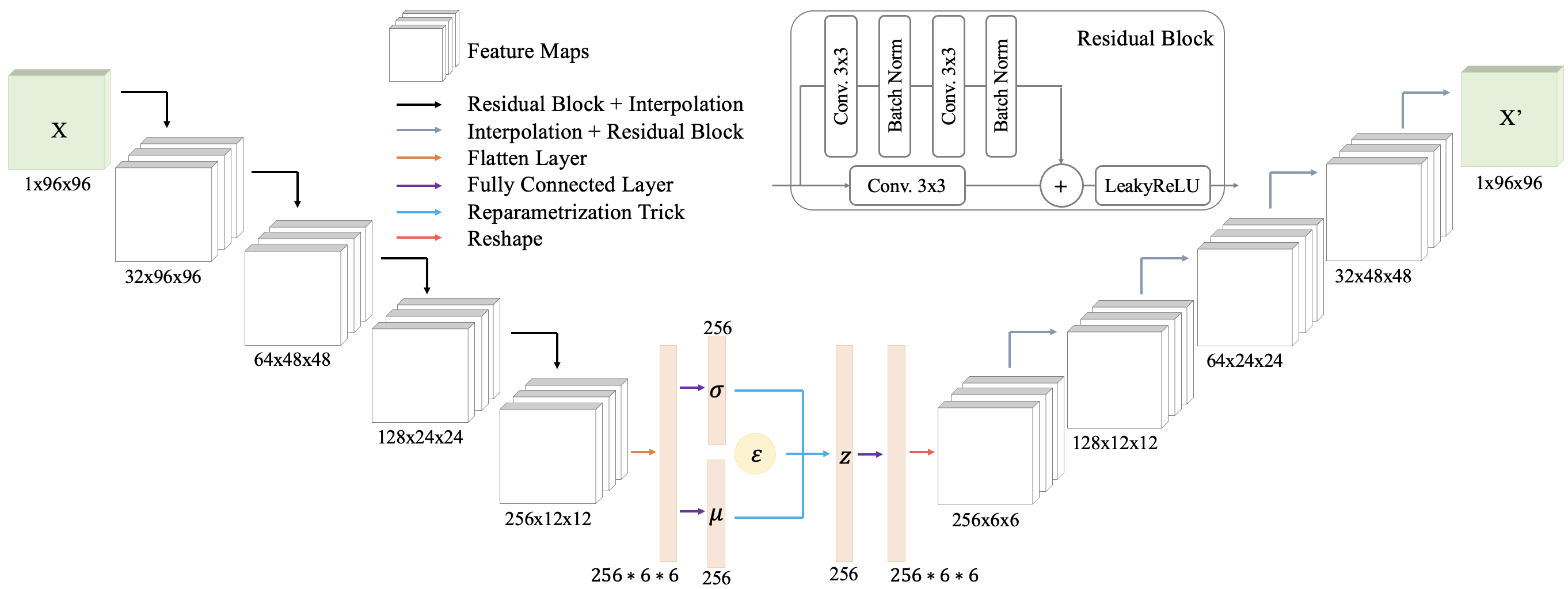}}
\caption{Residual variational autoencoder architecture with four residual blocks in each encoder and decoder, with a latent space dimension of $256$.}
\label{ResVAE}
\end{figure}

The proposed model is a Residual Variational Autoencoder (ResVAE), illustrated in Fig.~\ref{ResVAE}, which integrates residual blocks within a VAE framework to improve anatomical feature preservation. The encoder and decoder each comprise four residual blocks with channel dimensions of $32$, $64$, $128$, and $256$ (mirrored in the decoder). Each block consists of convolutional layers with batch normalization, residual connections, and Leaky ReLU activation. The encoder maps each 2D MRI slice to a $256$-dimensional latent space, parameterized by $\mu(x)$ and $\sigma(x)$, with latent sampling performed via reparameterization. This dimensionality balances representational capacity for complex pelvic anatomy with regularization to discourage memorization of pathological patterns. The decoder reconstructs the input using bilinear upsampling with a scale factor of $2$.

Training is performed for $100$ epochs using the AdamW optimizer (learning rate $1\times10^{-4}$), with early stopping (patience $15$) and gradient clipping (maximum norm $1.0$). The loss function combines pixel-level, structural, latent-space, and perceptual constraints, including MSE, SSIM, KL divergence with annealed $\beta$ (increased from $1\times10^{-5}$ to $1\times10^{-4}$ over the first $100$ epochs), and a perceptual loss computed using a ResNet-50 pretrained on RadImageNet~\cite{mei2022radimagenet}. Perceptual features are extracted from the first two residual stages of the network, weighted by $0.3$ and $0.15$, respectively. The overall training objective is defined as:
\begin{equation}
\begin{aligned}
\text{Total Loss\textsubscript{UAD}} = & \, \text{MSE Loss} + 0.5 \times \text{SSIM Loss} \\
&  + \beta(\text{epoch}) \times \text{KL Divergence} \\
& + 0.3 \times \text{Perceptual Loss\textsubscript{RadImageNet}}
\end{aligned}
\end{equation}

\subsubsection{Post-Processing} During inference, scans from D\textsubscript{\textit{Unhealthy}} are processed by the trained ResVAE to generate reconstructions. A residual image is computed as the difference between input and reconstruction, highlighting regions of abnormal deviation. To improve interpretability, the residual map is post-processed by applying a threshold at the $20$\textsubscript{th} percentile, selected empirically to maximize AUC, thereby suppressing low-magnitude errors. To emphasize anatomically relevant regions, a smooth radial weighting mask centered on the uterus is applied, defined by a Gaussian-like function of the Euclidean distance from the image center:
\begin{equation}
    \text{\textit{d}(center\textsubscript{x}, center\textsubscript{y})} = \sqrt{(x - \text{center\textsubscript{x}})^2 + (y - \text{center\textsubscript{y}})^2}
\end{equation}
\begin{equation}
    \text{Mask} = \exp\left(-\frac{\text{\textit{d}(center\textsubscript{x}, center\textsubscript{y})}^2}{2 \times \text{\textit{r}}^2}\right)
\end{equation}
where $r$ is set to $30$ pixels, as determined through empirical evaluation. Finally, the residual map is squared and smoothed using a median filter (kernel size $5$) to reduce noise and enhance regions of high reconstruction error.

\subsubsection{UAD Performance Analysis} 
The anomaly detection pipeline is evaluated under two complementary settings. First, quantitative benchmarking is performed on D\textsubscript{\textit{Unhealthy(UMD)}} to provide a standardized performance assessment, including an ablation study on the impact of synthetic data augmentation by comparing models trained with and without synthetic samples under identical conditions. The evaluation uses GT segmentations and standard metrics (Accuracy, Precision, Sensitivity, Specificity), with localization assessed via reconstruction error maps. ROC curves are computed and AUC is reported as a threshold-independent measure. Second, the model is applied to D\textsubscript{\textit{Unhealthy(in-house)}} to assess the impact of inter-observer variability by comparing identical predictions against annotations from multiple observers. This further enables evaluation across additional pathologies, including endometrial cancer, endometriosis, and adenomyosis.

\section{Results}
\label{sec:Results}
The evaluation of the proposed UAD framework includes an ablation study to quantify the effects of synthetic data augmentation and an analysis of interobserver variability.  In addition, the generated synthetic images are evaluated qualitatively.
\subsection{Data Set}
\label{sec:ResData}
The data sets exhibit substantial anatomical and technical imbalances. Uterine positioning is heavily skewed toward anteflexed (AF) configurations across all data sets: $88.09\%$ in the training set ($182$ anteverted (AV), $77$ retroverted (RV)), $68.26\%$ in D\textsubscript{\textit{Unhealthy(UMD)}} ($74$ AV, $68$ RV), and $84.85\%$ in the in-house subset. Retroflexed (RF) cases are markedly underrepresented, with only $35$ scans in training data ($14$ AV, $21$ RV), $66$ in D\textsubscript{\textit{Unhealthy(UMD)}} ($46$ AV, $20$ RV), and $5$ in the in-house subset ($3$ AV, $2$ RV). Field-strength distributions are similarly imbalanced. While D\textsubscript{\textit{Healthy}} comprises $56.46\%$ high-field $3.0~\mathrm{T}$ scans (alongside $60$ at $1.5~\mathrm{T}$ and $68$ at $0.55~\mathrm{T}$), D\textsubscript{\textit{Unhealthy(UMD)}} was acquired exclusively at $3.0~\mathrm{T}$. The in-house subset exhibits a different pattern with predominantly lower field strengths ($2$, $21$, and $10$ scans at $3.0~\mathrm{T}$, $1.5~\mathrm{T}$, and $0.55~\mathrm{T}$, respectively). 

\subsection{Synthetic Data Generation}

\begin{figure}[!t]
\centering
\includegraphics[width=0.63\columnwidth]{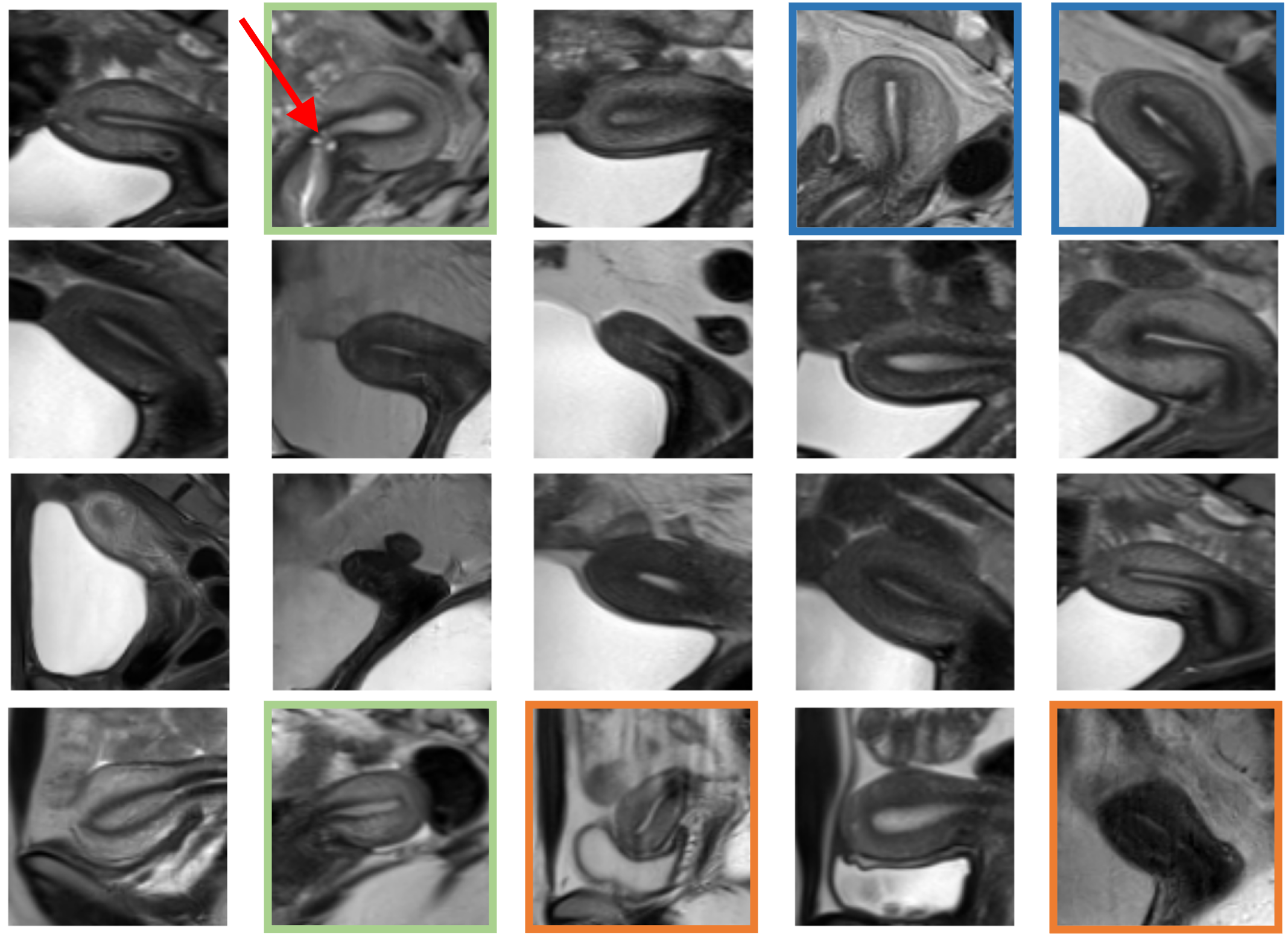}
\caption{Random synthetic samples illustrating uterine orientations (blue: RV, AF, green: RV, RF), image artifacts (orange), and Nabothian cysts (red arrows).}
\label{fig:SamplingGeneratedData}
\end{figure}

Random samples from D\textsubscript{\textit{Healthy\_Gen}} demonstrate substantial variability in the generated data. As depicted in Fig.~\ref{fig:SamplingGeneratedData}, the AV, AF uteri position is most frequently synthesized, while RV, AF (blue borders) and RV, RF (green borders) uteri are also represented. The absence of AV, RF position reflects the corresponding imbalance in the real training data (Sec.~\ref{sec:ResData}).
The generated images further capture both anatomical and acquisition-related variability. Orange borders indicate simulated artifacts, and the red arrow denotes the presence of Nabothian cysts. The samples exhibit diverse contrast profiles, including low uterine–background contrast and markedly higher contrast with reduced uterine signal intensity. Variations in myometrial signal intensity are also observed, as well as differences in uterine size, with one example (last row, second from right) demonstrating a markedly enlarged uterus, highlighting the diversity represented by the synthetic data.

\subsection{Unsupervised Anomaly Detection Performance}
\label{sec:UAD}
\subsubsection{Ablation Study} Results on D\textsubscript{\textit{Unhealthy(UMD)}} (Tab.~\ref{tab:umd_pathology_specific_results}) demonstrate that synthetic data augmentation consistently improves model performance across multiple pathologies. The largest gains are observed for Nabothian cysts, with accuracy increasing from $0.732$ to $0.765$ and AUC from $0.791$ to $0.826$. For uterine myomas, synthetic augmentation mainly enhances sensitivity from $0.708$ to $0.740$ and improves the AUC from $0.627$ to $0.646$.
%on UMD data
\begin{table*}[!t]
\centering
\small
\caption{
Performance metrics and AUC for pathology detection on D\textsubscript{\textit{Unhealthy(UMD)}} for models trained with and without synthetically generated data. Thresholds come from ROC curves. Best overall per-metric values are highlighted in bold.}
\begin{tabular}{lcccccccccc}
\multicolumn{1}{c}{} 
    & \multicolumn{2}{c}{\textbf{Accuracy$\uparrow$ }} 
    & \multicolumn{2}{c}{\textbf{Precision$\uparrow$ }} 
    & \multicolumn{2}{c}{\textbf{Sensitivity$\uparrow$ }} 
    & \multicolumn{2}{c}{\textbf{Specificity$\uparrow$ }} 
    & \multicolumn{2}{c}{\textbf{AUC$\uparrow$ }} \\
\cmidrule(lr){2-3} \cmidrule(lr){4-5} \cmidrule(lr){6-7} \cmidrule(lr){8-9} \cmidrule(lr){10-11}
Training data
    & w/ & w/o
    & w/ & w/o
    & w/ & w/o
    & w/ & w/o
    & w/ & w/o \\
\cmidrule(lr){2-2} \cmidrule(lr){3-3} 
\cmidrule(lr){4-4} \cmidrule(lr){5-5} 
\cmidrule(lr){6-6} \cmidrule(lr){7-7} 
\cmidrule(lr){8-8} \cmidrule(lr){9-9} 
\cmidrule(lr){10-10} \cmidrule(lr){11-11}
Uterine Myoma         
    & 0.618 & 0.625
    & 0.067 & 0.068
    & 0.740 & 0.708
    & 0.619 & 0.629
    & 0.646 & 0.627 \\
Nabothian Cyst        
    & 0.765 & 0.732
    & 0.035 & 0.021
    & 0.936 & 0.895
    & 0.765 & 0.732
    & 0.826 & 0.791 \\
\cmidrule(lr){2-2} \cmidrule(lr){3-3} 
\cmidrule(lr){4-4} \cmidrule(lr){5-5} 
\cmidrule(lr){6-6} \cmidrule(lr){7-7} 
\cmidrule(lr){8-8} \cmidrule(lr){9-9} 
\cmidrule(lr){10-10} \cmidrule(lr){11-11}
\textbf{Overall average} 
    & \textbf{0.692} & 0.679
    & \textbf{0.051} & 0.045
    & \textbf{0.838} & 0.802
    & \textbf{0.692} & 0.681
    & \textbf{0.736} & 0.709 \\
\end{tabular}
\label{tab:umd_pathology_specific_results}
\end{table*}

\subsubsection{Pathologies} 
\begin{figure}[t!]
    \centerline{\includegraphics[width=0.6\columnwidth]{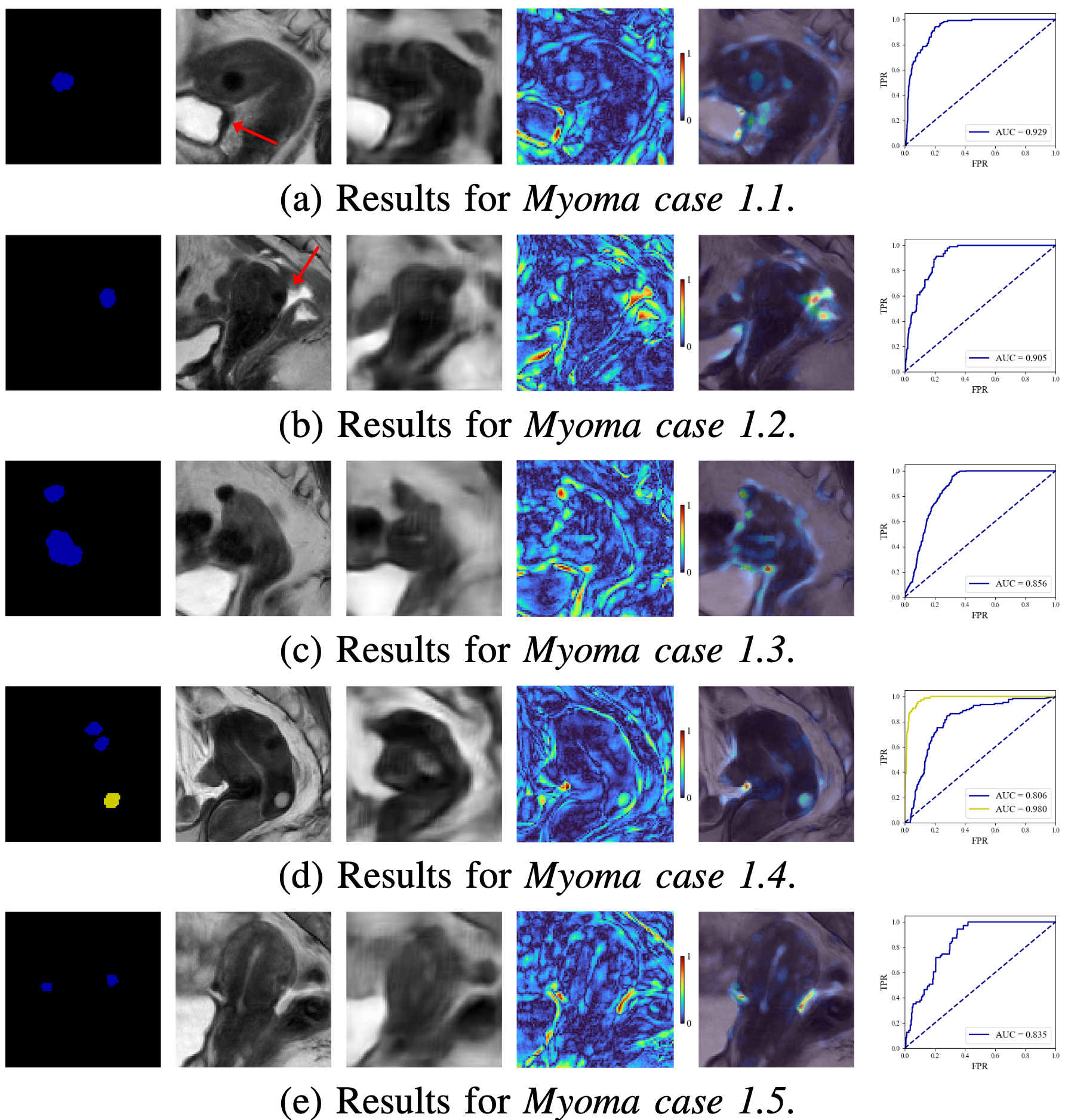}}
    \caption{Example cases from D\textsubscript{\textit{Unhealthy(UMD)}} including uterine myomas (blue) and Nabothian cysts (yellow). Original images, reconstructions, and subtraction maps are displayed as heat maps with an error range of $[0,1]$. Post-processed difference maps are overlaid on the original images with $50\%$ transparency, and corresponding ROC curves are presented on the right, color-coded by pathology.}
    \label{fig:PathoUMD}
\end{figure}

The following results demonstrate the pathology-specific performance of the model trained with synthetic data augmentation. As summarized in Tab.~\ref{tab:umd_pathology_specific_results}, Nabothian cyst detection achieves the highest scores for most evaluation metrics. However, despite high sensitivity and AUC, precision remains comparatively low ($0.35$). In contrast, uterine myomas indicate higher precision ($0.67$), albeit with lower overall sensitivity and AUC values. Fig.~\ref{fig:PathoUMD} depicts representative cases from the D\textsubscript{\textit{Unhealthy(UMD)}}, including uterine myomas and Nabothian cysts. Each example illustrates the corresponding GT segmentation, the original image, the reconstruction, the subtraction map, and the post-processed error overlay, as well as the associated ROC curve. Cases from the UMD and in-house cohorts are denoted as 1.x and 2.x, respectively.

In several myoma cases (cases 1.1 and 1.2), high AUC values (0.929 and 0.905, respectively) are achieved, while at the same time localized reconstruction errors occur that extend beyond the annotated lesion boundaries, particularly in the area near adjacent pelvic structures, which are highlighted with a red arrow. These areas reflect potential abnormal tissue that is not captured by the segmentations. Case 1.3 indicates reduced performance (AUC $0.856$), which is associated with a more anteverted uterine orientation in the reconstruction image, resulting in spatially distributed reconstruction error. In case 1.5, small uterine myomas demonstrate heterogeneous reconstruction errors, with the anterior myoma more clearly detected than the posterior one, which exhibits minimal contrast changes to surrounding tissue. Increased reconstruction error is additionally observed in adjacent brighter tissue posterior to the uterus.

\subsubsection{Uterine Positions} The model’s ability to detect pathologies across different uterine positions on D\textsubscript{\textit{Unhealthy(UMD)}} is depicted in Tab.~\ref{tab:uterine-position-performance}, reporting performance metrics averaged over all segmentations.
\begin{table}[!t]
\centering
\scriptsize
\caption{
Model performance across uterine positions (AV: anteverted, RV: retroverted, AF: anteflexed, RF: retroflexed) on D\textsubscript{\textit{Unhealthy(UMD)}}, averaged across all segmentations. Best per-metric values are highlighted in bold.}
\begin{tabular}{lccccc}
%\toprule
Position & \textbf{Accuracy}$\uparrow$ & \textbf{Precision}$\uparrow$ & \textbf{Sensitivity}$\uparrow$ & \textbf{Specificity}$\uparrow$ & \textbf{AUC}$\uparrow$ \\
\cmidrule(lr){2-2} \cmidrule(lr){3-3} \cmidrule(lr){4-4} \cmidrule(lr){5-5} \cmidrule(lr){6-6}
AF, AV       & $0.645$ & $\textbf{0.066}$ & $0.757$ & $0.640$ & $0.665$ \\
RF, AV      & $0.624$ & $0.063$ & $0.771$ & $0.626$ & $0.655$ \\
AF, RV      & $0.640$ & $0.060$ & $\textbf{0.811}$ & $0.641$ & $\textbf{0.699}$ \\
RF, RV     & $\textbf{0.665}$ & $0.053$ & $0.694$ & $\textbf{0.686}$ & $0.669$ \\
\end{tabular}
\label{tab:uterine-position-performance}
\end{table}
Performance remains comparable across positions, with AUC values ranging from $0.655$ to $0.699$. The highest sensitivity ($0.811$) and AUC ($0.699$) are observed for the AF, RV position, while the RF, RV cases achieve the highest accuracy ($0.665$) and specificity ($0.686$). In contrast, the RF, AV orientation exhibits the lowest accuracy, specificity, and AUC among all.

\subsubsection{Run-Time Evaluation} Real-time applicability was assessed by measuring inference reconstruction time of the VAE-based UAD pipeline. Experiments were conducted on an Apple M3 GPU (10-core, Metal Performance Shaders). The model achieved a reconstruction time of $10.8$ ms per slice, equivalent to $0.324$ s per 30-slice volume ($\approx$92.6 FPS).

\subsection{Inter-Observer Variability and Model Performance} 
\label{sec:Anno} After benchmarking on UMD, evaluation on the in-house cohort reveals that inter-observer variability profoundly impacts both lesion characterization and model performance assessment. Tab.~\ref{tab:compareAnnotationSize} summarizes mean lesion volumes with corresponding standard deviations and the number of annotated cases for each pathology.  
\begin{table}[!t]
\centering
\scriptsize
\caption{Mean lesion volumes (in mL $\pm$ stdv) for each pathology in D\textsubscript{\textit{Unhealthy(in-house)}}, reported separately for two experienced radiologists and one unexperienced observer (UO), illustrating inter-observer variability. The number of annotated cases is given in brackets.}
\setlength{\tabcolsep}{2pt}
\begin{tabular}{lccc}
Lesion type & \textbf{Radiologist 1} & \textbf{Radiologist 2} & \textbf{UO} \\
\cmidrule(lr){2-2} \cmidrule(lr){3-3} \cmidrule(lr){4-4}
Uterine Myoma      & 78.97 \textsubscript{$\pm$ 88.19} \,(16) & 69.23 \textsubscript{$\pm$ 84.03} \,(14) & 73.83 \textsubscript{$\pm$ 88.56} \,(11) \\
Nabothian Cyst      & 0.55 \textsubscript{$\pm$ 0.65} \,(16)   & 0.46 \textsubscript{$\pm$ 0.47} \,(14)   & 0.45 \textsubscript{$\pm$ 0.60} \,(12) \\
Endometrial Cancer  & 115.03 \textsubscript{$\pm$ 126.43} \,(5) & 66.72 \textsubscript{$\pm$ 74.38} \,(3) & 297.97 \textsubscript{$\pm$ 240.35} \,(5) \\
Endometriosis       & 21.62 \textsubscript{$\pm$ 17.46} \,(3)  & 7.74 \textsubscript{$\pm$ 0.00} \,(1)    & 36.38 \textsubscript{$\pm$ 48.37} \,(3) \\
Adenomyosis         & 4.95 \textsubscript{$\pm$ 7.64} \,(7)    & 14.72 \textsubscript{$\pm$ 7.31} \,(5)   & 15.93 \textsubscript{$\pm$ 12.09} \,(7) \\
\end{tabular}
\label{tab:compareAnnotationSize}
\end{table}
Annotation consistency correlates with lesion morphology. Focal lesions exhibit minimal volumetric variation (myomas: $69.23$~-~$78.97~\mathrm{mL}$, Nabothian cysts: $0.45$~-~$0.55~\mathrm{mL}$), while diffuse pathologies show volume ratios up to $4.5:1$ (endometrial cancer: $66.72$~-~$297.97~\mathrm{mL}$, endometriosis: $7.74$~-~$36.38~\mathrm{mL}$). Adenomyosis demonstrates pronounced variability both in lesion extent and case inclusion, with Radiologist~1 reporting markedly lower mean volumes ($4.95~\mathrm{mL}$) than the other annotators ($>14~\mathrm{mL}$). Beyond volumetric disagreement, annotator disagreement also affected case inclusion and pathology categorization, with some cases labeled healthy by Radiologist~1 but pathological by others, as well as inconsistencies between endometrial and cervical cancer classifications and between myoma and adenomyosis annotations.

\begin{figure}[!t]
\centerline{\includegraphics[width=0.6\columnwidth]{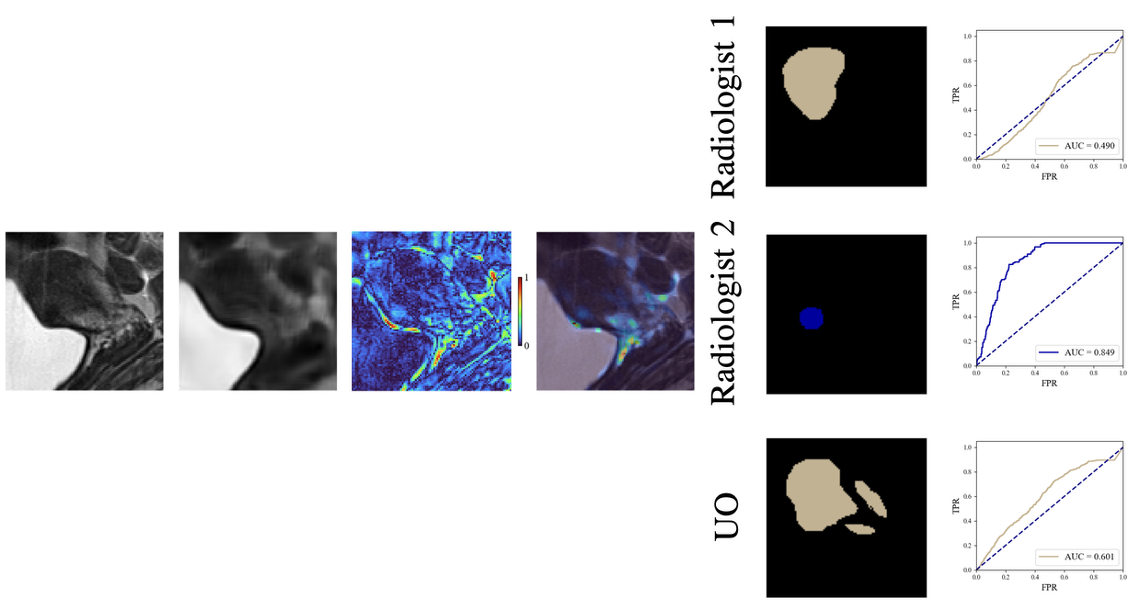}}
\caption{Example from D\textsubscript{\textit{Unhealthy(in-house)}} illustrating reconstruction-based anomaly localization with annotations from two expert radiologists and one inexperienced observer (UO). Blue denotes uterine myoma and brown adenomyosis. Corresponding ROC curves are included for each annotator.}
\label{fig:AnnotationDiff}
\end{figure}

Fig.~\ref{fig:AnnotationDiff} exemplifies these categorical inconsistencies, where Radiologist~2 identified a myoma while the others annotated the same case as adenomyosis. Evaluating a single reconstruction (left panel) against these divergent GT segmentations (right panels) yields AUC values spanning $0.490$ (adenomyosis, Radiologist~1) to $0.849$ (myoma, Radiologist~2), directly demonstrating how annotation variability translates to substantial performance assessment differences. Notably, the highest reconstruction error localizes to the bladder--uterus interface, a region not included in pathological annotations.

\begin{figure}[!t]
    \centerline{\includegraphics[width=0.6\columnwidth]{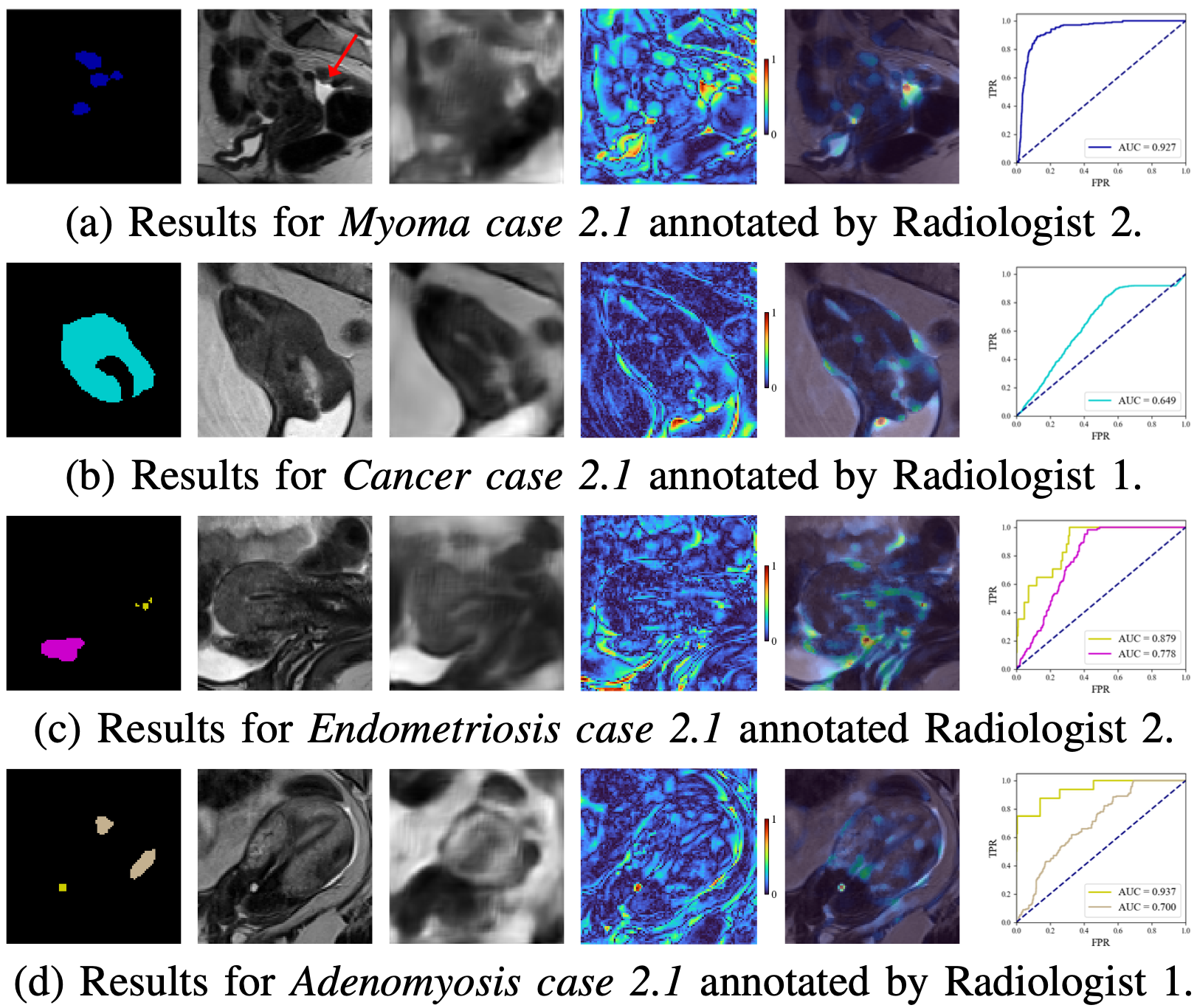}}
    \caption{Example cases from D\textsubscript{\textit{Unhealthy(in-house)}} including uterine myomas (blue), Nabothian cysts (yellow), cancer (cyan), endometriosis (magenta) and adenomyosis (brown). Original images, reconstructions, and subtraction maps are displayed as heat maps with an error range of $[0,1]$. Post-processed difference maps are overlaid on the original images with $50\%$ transparency, and corresponding ROC curves are presented on the right, color-coded by pathology.}
    \label{fig:Patho}
\end{figure}

Tab.~\ref{tab:evaluation-experienced-vs-unexperienced-sidebyside} compares performance metrics between experienced radiologists (mean) and an unexperienced observer, revealing substantial differences in the evaluated model performance.
\begin{table*}[!t]
\centering
\small
\caption{Performance metrics and AUC for pathology detection in D\textsubscript{\textit{Unhealthy(in-house)}}, comparing experienced radiologists (mean) and an unexperienced observer. Thresholds are derived from ROC curves. Best and second-best values are highlighted in bold and underlined. Overall performance is reported as a weighted average.} 
\begin{tabular}{lcccccccccc}
\multicolumn{1}{c}{} 
    & \multicolumn{2}{c}{\textbf{Accuracy$\uparrow$ }} 
    & \multicolumn{2}{c}{\textbf{Precision$\uparrow$ }} 
    & \multicolumn{2}{c}{\textbf{Sensitivity$\uparrow$ }} 
    & \multicolumn{2}{c}{\textbf{Specificity$\uparrow$ }} 
    & \multicolumn{2}{c}{\textbf{AUC$\uparrow$ }} \\
\cmidrule(lr){2-3} \cmidrule(lr){4-5} \cmidrule(lr){6-7} \cmidrule(lr){8-9} \cmidrule(lr){10-11}
 Annotator type   & Exp & Non & Exp & Non & Exp & Non & Exp & Non & Exp & Non \\
\cmidrule(lr){2-2} \cmidrule(lr){3-3} 
\cmidrule(lr){4-4} \cmidrule(lr){5-5} 
\cmidrule(lr){6-6} \cmidrule(lr){7-7} 
\cmidrule(lr){8-8} \cmidrule(lr){9-9} 
\cmidrule(lr){10-10} \cmidrule(lr){11-11}
Uterine Myoma    
    & \underline{0.658} & 0.598 & \underline{0.098} & 0.082 & 0.654 & 0.717 & \underline{0.657} & 0.602 & 0.619 & 0.619 \\
Nabothian Cyst  
    & \textbf{0.800} & \textbf{0.775} & 0.064 & 0.025 & \textbf{0.922} & \textbf{0.898} & \textbf{0.798} & \textbf{0.775} & \textbf{0.881} & \textbf{0.868} \\
Endometrial Cancer 
    & 0.510 & 0.495 & \textbf{0.127} & \textbf{0.293} & 0.677 & 0.798 & 0.489 & 0.414 & 0.515 & 0.605 \\
Endometriosis   
    & 0.521 & \underline{0.729} & 0.056 & 0.080 & \underline{0.901} & \underline{0.837} & 0.509 & \underline{0.725} & \underline{0.683} & \underline{0.839} \\
Adenomyosis      
    & 0.565 & 0.573 & 0.082 & \underline{0.127} & 0.799 & 0.781 & 0.553 & 0.550 & 0.664 & 0.678 \\
\cmidrule(lr){2-2} \cmidrule(lr){3-3} 
\cmidrule(lr){4-4} \cmidrule(lr){5-5} 
\cmidrule(lr){6-6} \cmidrule(lr){7-7} 
\cmidrule(lr){8-8} \cmidrule(lr){9-9} 
\cmidrule(lr){10-10} \cmidrule(lr){11-11}
\textbf{Overall Average} 
    & 0.639 & 0.627 & 0.095 & 0.129 & 0.762 & 0.790 & 0.632 & 0.612 & 0.632 & 0.702 \\
\end{tabular}
\label{tab:evaluation-experienced-vs-unexperienced-sidebyside}
\end{table*}
Detection performance correlates strongly with lesion morphology (compare Fig.~\ref{fig:Patho} presenting a selection of one case per pathology). Focal pathologies with sharp contrast boundaries achieve highest performance, with Nabothian cysts (mean AUC $0.868$, e.g., $0.937$ in Fig.~\ref{fig:Patho}) and uterine myomas (mean AUC $0.619$, e.g., $0.927$) exhibiting localized reconstruction errors that align precisely with lesion boundaries. Conversely, diffuse pathologies demonstrate substantially lower detectability due to distributed, interface-aligned errors rather than focal responses. In edometrial cancer (mean AUC $0.515$, e.g., $0.649$) and adenomyosis (mean AUC $0.664$, e.g., $0.700$) cases, reconstruction errors are present along anatomical boundaries (endometrial layer, cervix--uterus junction, JZ), while endometriosis achieves intermediate performance (mean AUC $0.683$, e.g., $0.839$) depending on lesion localization. Cases with minimal signal intensity variation exhibit spatially diffuse reconstruction patterns, demonstrating model sensitivity to global rather than focal deviations. Annotation experience introduces systematic bias, with the unexperienced observer's annotations yielding higher overall AUC ($0.702$ versus $0.632$) driven primarily by broader lesion delineations that increase true positive rates at the cost of specificity.

\section{Discussion}
\label{sec:Disc}
\subsection{Summary}
This work establishes a foundation for unsupervised anomaly detection in female pelvic MRI. By adopting a disease- and parameter-independent approach that reflects the anatomical complexity and variability of the pelvis, it paves the way for future research and supports the development of AI-driven real-time workflows for pelvic MRI.

\subsection{Comparison to the State of the Art}
 Diagnostic performance of clinical radiological-led interpretation of pelvis MRI, while achieving overall accuracies of 80-90\% for benign conditions and above 90\% for malignant staging,~\cite{foti2018endometriosis, thomassin2020ovarian}, depends strongly on expertise and decreases in atypical or subtle cases~\cite{harth2023application}. In complex conditions such as deep infiltrating endometriosis, MRI alone is often insufficient~\cite{pascoal2022strengths}, contributing to delayed diagnosis and motivating automated real-time decision support.

To date, UAD has not been explored in female pelvic MRI. Prior work in anatomically more homogeneous regions demonstrates, however, the potential of generative models. Baur et al.~\cite{baur2021modeling, baur2019deep, baur2021autoencoders} reported high AUC values (e.g., $0.945$ for multiple sclerosis) using autoencoders and VAE-GANs in brain MRI, while Patsanis et al.~\cite{patsanis2023comparison} applied GAN-based UAD to prostate MRI, achieving moderate performance ($AUC = 0.76$) and reporting sensitivity to anatomical variability. These findings highlight the increasing difficulty of UAD in regions with complex and heterogeneous anatomy as present in the pelvis. Recent large-scale benchmarks, including BMAD~\cite{bao2024bmad} and MedAnomaly~\cite{cai2025medianomaly}, further demonstrate strong UAD performance across standardized tasks such as brain MRI, liver CT, and chest X-ray. However, these applications support organ-centric cropping, background suppression, or patch-based evaluation, limiting anatomical context. Such assumptions are less applicable to pelvic MRI, where clinically relevant abnormalities often arise from global anatomical alterations, inter-organ relationships, or diffuse tissue changes rather than isolated focal lesions. Consequently, direct benchmarking between these protocols is not methodologically comparable. Within this substantially more challenging setting, our fully unsupervised, disease- and parameter-agnostic approach achieves an overall AUC of $0.736$ on the UMD across heterogeneous pelvic pathologies, with strong performance for example for specific findings such as Nabothian cysts (AUC $= 0.826$). Importantly, these performance values should be interpreted as conservative estimates, as quantitative evaluation is inherently constrained by inter-observer variability and GT definitions limited by a high degree on subjectivity. These findings highlight both the complexity of pelvic anomaly detection and the need to contextualize quantitative metrics within anatomical variability and annotation uncertainty.

\subsection{Strengths and Limitations}
The proposed UAD pipeline demonstrated several strengths in the detection of female pelvic disease. Notably, it performed comparably across uterine positions, validating the concept of reconstructing healthy anatomy to detect deviations. The model performed particularly well for focal lesions, such as uterine myomas and Nabothian cysts, enabling clear anomaly localization. Another key strength lies in its robustness to imaging variability. Despite the absence of a standardized MRI protocol and substantial heterogeneity in the training data, including variations in field strength, resolution, contrast, and sequence type, the model generalized well to unseen data. This was evidenced by consistent performance across uterine myoma and Nabothian cyst cases in both the UMD and in-house data sets, indicating strong multi-center applicability. Such generalizability is essential for real-world deployment, where diverse acquisition protocols are common. To address limited data availability and increase training diversity, synthetically generated samples expanded the data set, further improving detection robustness, particularly for Nabothian cysts, which already exhibited stable performance prior to augmentation.

A key limitation of the proposed UAD pipeline is its dependence on high-quality and representative training data. Mislabeled or unhealthy cases may cause the model to reconstruct pathological patterns, reducing detection performance. Rare anatomical configurations, such as retroflexed uterine positions, are currently underrepresented, potentially limiting generalization and increasing reconstruction errors. This imbalance also propagates to synthetic data generation, as DDPMs tend to reproduce dominant anatomical patterns rather than compensate for distribution bias~\cite{dar2025unconditional}. Architecturally, the probabilistic nature of VAEs introduces output variability, affecting reproducibility and limiting sensitivity to subtle or low-contrast anomalies, including adenomyosis and early-stage malignancies. Moreover, the evaluation of the model is further limited by the quality and consistency of the GT annotations. Inter-observer disagreement arises not only in delineating lesion boundaries, but also in determining whether a case is pathological and in identifying the specific pathology. Especially in the case of adenomyosis, diagnostic discrepancies may stem from reliance on a single sequence, where the condition is often subtle and heavily influenced by the visibility of the JZ~\cite{Bazot2001, Novellas2011}. As a result, identical model predictions can yield significantly different performance metrics depending on the reference segmentation. This issue is well documented in medical imaging, where label noise arising from inter-observer variability and inconsistent annotations can significantly impact model performance~\cite{karimi2020deep}. In addition, ground-truth masks may omit pathological regions outside the uterine structures, leading to apparent false positives when the model correctly identifies unannotated abnormalities. Together, these findings highlight a fundamental challenge in pelvic MRI: while diagnostic ambiguity motivates the need for real-time decision support, reliable model development and evaluation depend critically on consistent, high-quality reference annotations, emphasizing the importance of standardized labeling, multi-reader consensus, and explicit reporting of inter-observer variability.

\subsection{Research and Clinical Implications}
The proposed UAD pipeline establishes a disease-agnostic baseline for pelvic anomaly detection on MRI data, enabling identification of diverse pathologies without labeled data with anomalies. Evaluation on the publicly available UMD provides a reproducible benchmark for future studies. Pelvic disease diagnosis is often hindered by anatomical variability, limited awareness of conditions such as endometriosis, and reliance on reader expertise. By highlighting deviations from learned healthy anatomy, the proposed approach offers an objective reference that may support earlier detection and reduce diagnostic delays. Recent work in prostate MRI further demonstrates that AI-based analysis can achieve expert-level or superior diagnostic performance, even without full clinical context~\cite{saha2024artificial}.

A key clinical application of the proposed UAD framework is its readiness for integration into real-time MRI, analogous to recent advances in fetal imaging, enabling adaptive acquisitions and immediate feedback during scanning~\cite{aviles2025real, verdera2025heron}. Such integration could optimize imaging protocols, reduce unnecessary interventions, and improve diagnostic efficiency and patient experience. As a disease-agnostic approach, the model may also support radiologist training by highlighting suspicious regions without relying on predefined pathology labels, thereby improving interpretability. Extending the framework to multimodal imaging with additional contrasts such as T1-weighted imaging could further enhance AI-guided pelvic diagnostics and support more equitable reproductive healthcare. In addition, sensitivity to subtle anatomical deviations may offer insights into disease development, particularly for relatively underexplored conditions such as endometriosis~\cite{anchan2024unveiling}. Notably, qualitative review of the unhealthy cohorts revealed regions with consistently high reconstruction error that were not included in the reference annotations (e.g., Fig.~\ref{fig:PathoUMD} a, b), suggesting potential omissions in the ground truth. If such regions correspond to unannotated pathology, quantitative metrics may underestimate true model performance, as the UAD framework is designed to detect any abnormal pelvic tissue rather than only predefined lesion labels.

\subsection{Future steps}
Expanding the data set to include broader demographic representation and rarer pelvic diseases is essential to improve model robustness and generalizability. Methodologically, integrating alternative architectures and learning paradigms may further enhance UAD performance. Incorporating domain-specific anatomical priors, such as typical uterine shape, size, and spatial configuration, could guide representation learning and improve discrimination between normal and pathological anatomy. In addition, contrastive and self-supervised learning strategies may increase sensitivity to subtle abnormalities by explicitly modeling intra-class anatomical variability without requiring pathological labels. Recent advances in diffusion-based anomaly detection, particularly demonstrated in brain MRI~\cite{behrendt2024patched,baugh2024image}, further highlight the potential of these approaches for future application to female pelvic imaging.

\section{Conclusion}
This work presents a deep generative framework for unsupervised anomaly detection in female pelvic MRI, designed to model the substantial anatomical variability of healthy pelvic anatomy. By learning the distribution of normal tissue, the proposed approach identifies abnormalities as reconstruction deviations, removing the need for annotated pathological data. The low-latency design supports integration into real-time MRI workflows, enabling immediate visual feedback during image acquisition. Such capability has the potential to support radiologists in identifying subtle abnormalities, reduce diagnostic uncertainty, and facilitate more personalized imaging strategies. Overall, this study establishes a benchmark for disease-agnostic anomaly detection in the female pelvis and provides a foundation for future research toward AI-assisted, real-time pelvic MRI.

\section*{Acknowledgments}
\noindent This work was supported by the High Tech Agenda of
the Free State of Bavaria, funding from the BayStMGP [EndoKI], DFG Heisenberg funding [502024488] and an ERC Starting grant EARTHWORM [101165242].
\bibliography{refs}

@article{gao2025rising,
  title = {Rising global burden of common gynecological diseases in women of childbearing age from 1990 to 2021: an update from the Global Burden of Disease Study 2021},
  volume = {22},
  ISSN = {1742-4755},
  DOI = {https://doi.org/10.1186/s12978-025-02013-1},
  number = {1},
  journal = {Reproductive Health},
  publisher = {Springer Science and Business Media LLC},
  author = {Gao,  Yidan and Wang,  Xuemei and Wang,  Qian and Jiang,  Lijuan and Wu,  Cuixiu and Guo,  Yuanshuo and Cui,  Na and Tang,  Haoneng and Tang,  Lingli},
  year = {2025},
  month = apr 
}

@article {Sebastianojnumed.124.267546,
  title = {Molecular Imaging in Gynecology: Beyond Cancer},
  volume = {65},
  ISSN = {2159-662X},
  DOI = {https://doi.org/10.2967/jnumed.124.267546},
  number = {7},
  journal = {Journal of Nuclear Medicine},
  publisher = {Society of Nuclear Medicine},
  author = {Sebastiano,  Joni and Rodriguez,  Cindy and Samuels,  Zachary V. and Pepin,  Kristen and Zeglis,  Brian M.},
  year = {2024},
  month = jun,
  pages = {998–1003}
}

@article{Bazot2001,
  title = {Ultrasonography compared with magnetic resonance imaging for the diagnosis of adenomyosis: correlation with histopathology},
  volume = {16},
  ISSN = {0268-1161},
  DOI = {https://doi.org/10.1093/humrep/16.11.2427},
  number = {11},
  journal = {Human Reproduction},
  publisher = {Oxford University Press (OUP)},
  author = {Bazot,  Marc and Cortez,  Annie and Darai,  Emile and Rouger,  Jérome and Chopier,  Jocelyne and Antoine,  Jean-Marie and Uzan,  Serge},
  year = {2001},
  month = nov,
  pages = {2427–2433}
}

@article{Novellas2011,
  title = {MRI Characteristics of the Uterine Junctional Zone: From Normal to the Diagnosis of Adenomyosis},
  volume = {196},
  ISSN = {1546-3141},
  DOI = {https://doi.org/10.2214/ajr.10.4877},
  number = {5},
  journal = {American Journal of Roentgenology},
  publisher = {American Roentgen Ray Society},
  author = {Novellas,  Sébastien and Chassang,  Madleen and Delotte,  Jerome and Toullalan,  Olivier and Chevallier,  Anne and Bouaziz,  Jerome and Chevallier,  Patrick},
  year = {2011},
  month = may,
  pages = {1206–1213}
}

@article{zabihollahy2021fully,
  title = {Fully automated multiorgan segmentation of female pelvic magnetic resonance images with coarse‐to‐fine convolutional neural network},
  volume = {48},
  ISSN = {2473-4209},
  DOI = {https://doi.org/10.1002/mp.15268},
  number = {11},
  journal = {Medical Physics},
  publisher = {Wiley},
  author = {Zabihollahy,  Fatemeh and Viswanathan,  Akila N and Schmidt,  Ehud J and Morcos,  Marc and Lee,  Junghoon},
  year = {2021},
  month = oct,
  pages = {7028–7042}
}

@article{satushe2025ai,
  title = {AI in MRI brain tumor diagnosis: A systematic review of machine learning and deep learning advances (2010–2025)},
  volume = {263},
  ISSN = {0169-7439},
  DOI = {https://doi.org/10.1016/j.chemolab.2025.105414},
  journal = {Chemometrics and Intelligent Laboratory Systems},
  publisher = {Elsevier BV},
  author = {Satushe,  Vaidehi and Vyas,  Vibha and Metkar,  Shilpa and Singh,  Davinder Paul},
  year = {2025},
  month = aug,
  pages = {105414}
}

@article{dungate2024assessing,
  title = {Assessing the Utility of artificial intelligence in endometriosis: Promises and pitfalls},
  volume = {20},
  ISSN = {1745-5065},
  DOI = {https://doi.org/10.1177/17455057241248121},
  journal = {Women’s Health},
  publisher = {SAGE Publications},
  author = {Dungate,  Brie and Tucker,  Dwayne R and Goodwin,  Emma and Yong,  Paul J},
  year = {2024},
  month = jan 
}

@article{aftab2025artificial,
  title = {Artificial Intelligence in Obstetrics and Gynaecology: Advancing Precision and Personalised Care},
  ISSN = {2168-8184},
  DOI = {https://doi.org/10.7759/cureus.86929},
  journal = {Cureus},
  publisher = {Springer Science and Business Media LLC},
  author = {Aftab,  Nida},
  year = {2025},
  month = jun 
}

@article{verdera2025heron,
  title = {HERON: High-Efficiency Real-Time mOtion quantification and re-acquisitioN for Fetal diffusion MRI},
  ISSN = {1558-254X},
  DOI = {https://doi.org/10.1109/tmi.2025.3569853},
  journal = {IEEE Transactions on Medical Imaging},
  publisher = {Institute of Electrical and Electronics Engineers (IEEE)},
  author = {Verdera,  Jordina Aviles and Bortolazzi,  Antonia and Silva,  Sara Neves and Payette,  Kelly and Clair,  Kamilah St. and McElroy,  Sarah and Malik,  Shaihan and Hajnal,  Joseph and Tomi-Tricot,  Raphael and Rutherford,  Mary and Hutter,  Jana},
  year = {2025},
  pages = {1–1}
}

@article{aviles2025real,
  title = {Real‐time fetal brain and placental T2* mapping at 0.55T MRI},
  volume = {94},
  ISSN = {1522-2594},
  DOI = {https://doi.org/10.1002/mrm.30497},
  number = {2},
  journal = {Magnetic Resonance in Medicine},
  publisher = {Wiley},
  author = {Aviles Verdera,  Jordina and Neves Silva,  Sara and Payette,  Kelly M. and Tomi‐Tricot,  Raphael and Hall,  Megan and Story,  Lisa and Malik,  Shaihan J. and Hajnal,  Joseph V. and Rutherford,  Mary A. and Hutter,  Jana},
  year = {2025},
  month = mar,
  pages = {615–624}
}

@article{nougaret2022mri,
  title = {MRI in female pelvis: an ESUR/ESR survey},
  volume = {13},
  ISSN = {1869-4101},
  DOI = {https://doi.org/10.1186/s13244-021-01152-w},
  number = {1},
  journal = {Insights into Imaging},
  publisher = {Springer Science and Business Media LLC},
  author = {Nougaret,  Stephanie and Lakhman,  Yulia and Gourgou,  Sophie and Kubik-Huch,  Rahel and Derchi,  Lorenzo and Sala,  Evis and Forstner,  Rosemarie},
  year = {2022},
  month = mar 
}

@article{sudderuddin2014mri,
  title = {MRI appearances of benign uterine disease},
  volume = {69},
  ISSN = {0009-9260},
  DOI = {https://doi.org/10.1016/j.crad.2014.05.108},
  number = {11},
  journal = {Clinical Radiology},
  publisher = {Elsevier BV},
  author = {Sudderuddin,  S. and Helbren,  E. and Telesca,  M. and Williamson,  R. and Rockall,  A.},
  year = {2014},
  month = nov,
  pages = {1095–1104}
}

@article{hudelist2012diagnostic,
  title = {Diagnostic delay for endometriosis in Austria and Germany: causes and possible consequences},
  volume = {27},
  ISSN = {1460-2350},
  DOI = {https://doi.org/10.1093/humrep/des316},
  number = {12},
  journal = {Human Reproduction},
  publisher = {Oxford University Press (OUP)},
  author = {Hudelist,  G. and Fritzer,  N. and Thomas,  A. and Niehues,  C. and Oppelt,  P. and Haas,  D. and Tammaa,  A. and Salzer,  H.},
  year = {2012},
  month = sep,
  pages = {3412–3416}
}

@article{bulun2019,
  title = {Endometriosis},
  volume = {40},
  ISSN = {1945-7189},
  DOI = {https://doi.org/10.1210/er.2018-00242},
  number = {4},
  journal = {Endocrine Reviews},
  publisher = {The Endocrine Society},
  author = {Bulun,  Serdar E and Yilmaz,  Bahar D and Sison,  Christia and Miyazaki,  Kaoru and Bernardi,  Lia and Liu,  Shimeng and Kohlmeier,  Amanda and Yin,  Ping and Milad,  Magdy and Wei,  JianJun},
  year = {2019},
  month = apr,
  pages = {1048–1079}
}

@article{cai2024artificial,
  title = {Artificial intelligence in abdominal and pelvic ultrasound imaging: current applications},
  volume = {50},
  ISSN = {2366-0058},
  DOI = {https://doi.org/10.1007/s00261-024-04640-x},
  number = {4},
  journal = {Abdominal Radiology},
  publisher = {Springer Science and Business Media LLC},
  author = {Cai,  Lie and Pfob,  André},
  year = {2024},
  month = nov,
  pages = {1775–1789}
}

@article{baur2021modeling,
  title = {Modeling Healthy Anatomy with Artificial Intelligence for                     Unsupervised Anomaly Detection in Brain MRI},
  volume = {3},
  ISSN = {2638-6100},
  DOI = {https://doi.org/10.1148/ryai.2021190169},
  number = {3},
  journal = {Radiology: Artificial Intelligence},
  publisher = {Radiological Society of North America (RSNA)},
  author = {Baur,  Christoph and Wiestler,  Benedikt and Muehlau,  Mark and Zimmer,  Claus and Navab,  Nassir and Albarqouni,  Shadi},
  year = {2021},
  month = may,
  pages = {e190169}
}

@article{baur2021autoencoders,
  title = {Autoencoders for unsupervised anomaly segmentation in brain MR images: A comparative study},
  volume = {69},
  ISSN = {1361-8415},
  DOI = {https://doi.org/10.1016/j.media.2020.101952},
  journal = {Medical Image Analysis},
  publisher = {Elsevier BV},
  author = {Baur,  Christoph and Denner,  Stefan and Wiestler,  Benedikt and Navab,  Nassir and Albarqouni,  Shadi},
  year = {2021},
  month = apr,
  pages = {101952}
}

@article{raad2023unsupervised,
  title = {Unsupervised abnormality detection in neonatal MRI brain scans using deep learning},
  volume = {13},
  ISSN = {2045-2322},
  DOI = {https://doi.org/10.1038/s41598-023-38430-0},
  number = {1},
  journal = {Scientific Reports},
  publisher = {Springer Science and Business Media LLC},
  author = {Raad,  Jad Dino and Chinnam,  Ratna Babu and Arslanturk,  Suzan and Tan,  Sidhartha and Jeong,  Jeong-Won and Mody,  Swati},
  year = {2023},
  month = jul 
}

@inproceedings{behrendt2024patched,
  booktitle={Medical Imaging with Deep Learning},
  DOI = {https://doi.org/10.48550/ARXIV.2303.03758},
  author = {Behrendt,  Finn and Bhattacharya,  Debayan and Kr\"{u}ger,  Julia and Opfer,  Roland and Schlaefer,  Alexander},
  title = {Patched Diffusion Models for Unsupervised Anomaly Detection in Brain MRI},
  publisher = {arXiv},
pages={1019--1032},
  year = {2023},
  copyright = {arXiv.org perpetual,  non-exclusive license},
organization={PMLR}
}

@article{patsanis2023comparison,
  title = {A comparison of Generative Adversarial Networks for automated prostate cancer detection on T2-weighted MRI},
  volume = {39},
  ISSN = {2352-9148},
  DOI = {https://doi.org/10.1016/j.imu.2023.101234},
  journal = {Informatics in Medicine Unlocked},
  publisher = {Elsevier BV},
  author = {Patsanis,  Alexandros and Sunoqrot,  Mohammed R.S. and Langørgen,  Sverre and Wang,  Hao and Selnæs,  Kirsten M. and Bertilsson,  Helena and Bathen,  Tone F. and Elschot,  Mattijs},
  year = {2023},
  pages = {101234}
}

@article{figueredo2024automatic,
  title = {Automatic segmentation of deep endometriosis in the rectosigmoid using deep learning},
  volume = {151},
  ISSN = {0262-8856},
  DOI = {https://doi.org/10.1016/j.imavis.2024.105261},
  journal = {Image and Vision Computing},
  publisher = {Elsevier BV},
  author = {Figueredo,  Weslley Kelson Ribeiro and Silva,  Aristófanes Corr\^ea and de Paiva,  Anselmo Cardoso and Diniz,  João Otávio Bandeira and Brandão,  Alice and Oliveira,  Marco Aurelio Pinho},
  year = {2024},
  month = nov,
  pages = {105261}
}

@article{saha2024artificial,
  title={Artificial intelligence and radiologists in prostate cancer detection on MRI (PI-CAI): an international, paired, non-inferiority, confirmatory study},
  author={Saha, Anindo and Bosma, Joeran S and Twilt, Jasper J and van Ginneken, Bram and Bjartell, Anders and Padhani, Anwar R and Bonekamp, David and Villeirs, Geert and Salomon, Georg and Giannarini, Gianluca and others},
  volume = {25},
  ISSN = {1470-2045},
  DOI = {https://doi.org/10.1016/s1470-2045(24)00220-1},
  number = {7},
  journal = {The Lancet Oncology},
  publisher = {Elsevier BV},
  year = {2024},
  month = jul,
  pages = {879–887}
}

@article{cai2025artificial,
  title = {Artificial intelligence in abdominal and pelvic ultrasound imaging: current applications},
  volume = {50},
  ISSN = {2366-0058},
  DOI = {https://doi.org/10.1007/s00261-024-04640-x},
  number = {4},
  journal = {Abdominal Radiology},
  publisher = {Springer Science and Business Media LLC},
  author = {Cai,  Lie and Pfob,  André},
  year = {2024},
  month = nov,
  pages = {1775–1789}
}

@article{tong2024best,
  title = {Best Practices: Ultrasound Versus MRI in the Assessment of Pelvic Endometriosis},
  volume = {223},
  ISSN = {1546-3141},
  DOI = {https://doi.org/10.2214/ajr.24.31085},
  number = {6},
  journal = {American Journal of Roentgenology},
  publisher = {American Roentgen Ray Society},
  author = {Tong,  Angela and Cope,  Adela G. and Waters,  Timothy L. and McDonald,  Jennifer S. and VanBuren,  Wendaline M.},
  year = {2024},
  month = dec 
}

@article{mahmoud2025comparative,
  title = {Comparative study between MRI and Ultrasound in evaluation of Uterine lesions},
  volume = {0},
  ISSN = {2357-0016},
  DOI = {https://doi.org/10.21608/bmfj.2025.344381.2286},
  number = {0},
  journal = {Benha Medical Journal},
  publisher = {Egyptian Knowledge Bank},
  author = {El said,  Asmaa Yousef and Mahmoud,  Nadia Ali and Khater,  Hamada Mohamed},
  year = {2025},
  month = feb,
  pages = {0–0}
}

@article{pan2024large,
  title = {Large-scale uterine myoma MRI dataset covering all FIGO types with pixel-level annotations},
  volume = {11},
  ISSN = {2052-4463},
  DOI = {https://doi.org/10.1038/s41597-024-03170-x},
  number = {1},
  journal = {Scientific Data},
  publisher = {Springer Science and Business Media LLC},
  author = {Pan,  Haixia and Chen,  Minghuang and Bai,  Wenpei and Li,  Bin and Zhao,  Xiaoran and Zhang,  Meng and Zhang,  Dongdong and Li,  Yanan and Wang,  Hongqiang and Geng,  Haotian and Kong,  Weiya and Yin,  Cong and Han,  Linfeng and Lan,  Jiahua and Zhao,  Tian},
  year = {2024},
  month = apr 
}

@article{pan2023,
author = "Pan, Haixia and Chen, Minghuang and Bai, Wenpei and Li, Bin and Zhao, Xiaoran and Zhang, Meng and Zhang, Dongdong and Li, Yanan and Wang, Hongqiang",
title = "{UMD.zip}",
year = "2023",
month = "6",
journal = "figshare",
doi = "https://doi.org/10.6084/m9.figshare.23541312.v3"
}

@article{py06nimg,
  title = {User-guided 3D active contour segmentation of anatomical structures: Significantly improved efficiency and reliability},
  volume = {31},
  ISSN = {1053-8119},
  DOI = {https://doi.org/10.1016/j.neuroimage.2006.01.015},
  number = {3},
  journal = {NeuroImage},
  publisher = {Elsevier BV},
  author = {Yushkevich,  Paul A. and Piven,  Joseph and Hazlett,  Heather Cody and Smith,  Rachel Gimpel and Ho,  Sean and Gee,  James C. and Gerig,  Guido},
  year = {2006},
  month = jul,
  pages = {1116–1128}
}

@misc{dicomstandard,
  author       = {National Electrical Manufacturers Association},
  title        = {DICOM Standards},
  year         = {2021},
  url          = {https://www.dicomstandard.org},
  note         = {Accessed: 2025-03-03}
}

@misc{nifti,
  publisher       = {National Institute of Mental Health},
author = {Cox, RW and Ashburner, J and Breman, H and Fissell, K and Haselgrove, C and Holmes, CJ and Lancaster, JL and Rex, DE and Smith, SM and Woodward, JB and Strother, SC},
  title        = {NIfTI - Neuroimaging Informatics Technology Initiative},
  year         = {2011},
  url          = {https://nifti.nimh.nih.gov},
  note         = {Accessed: 2025-03-03}
}

@article{yaniv2018simpleitk,
  title = {SimpleITK Image-Analysis Notebooks: a Collaborative Environment for Education and Reproducible Research},
  volume = {31},
  ISSN = {1618-727X},
  DOI = {https://doi.org/10.1007/s10278-017-0037-8},
  number = {3},
  journal = {Journal of Digital Imaging},
  publisher = {Springer Science and Business Media LLC},
  author = {Yaniv,  Ziv and Lowekamp,  Bradley C. and Johnson,  Hans J. and Beare,  Richard},
  year = {2017},
  month = nov,
  pages = {290–303}
}

@article{reza2004realization,
  title = {Realization of the Contrast Limited Adaptive Histogram Equalization (CLAHE) for Real-Time Image Enhancement},
  volume = {38},
  ISSN = {0922-5773},
  DOI = {https://doi.org/10.1023/b:vlsi.0000028532.53893.82},
  number = {1},
  journal = {Journal of VLSI signal processing systems for signal,  image and video technology},
  publisher = {Springer Science and Business Media LLC},
  author = {Reza,  Ali M.},
  year = {2004},
  month = aug,
  pages = {35–44}
}

@article{foti2018endometriosis,
  title = {Endometriosis: clinical features,  MR imaging findings and pathologic correlation},
  volume = {9},
  ISSN = {1869-4101},
  DOI = {https://doi.org/10.1007/s13244-017-0591-0},
  number = {2},
  journal = {Insights into Imaging},
  publisher = {Springer Science and Business Media LLC},
  author = {Foti,  Pietro Valerio and Farina,  Renato and Palmucci,  Stefano and Vizzini,  Ilenia Anna Agata and Libertini,  Norma and Coronella,  Maria and Spadola,  Saveria and Caltabiano,  Rosario and Iraci,  Marco and Basile,  Antonio and Milone,  Pietro and Cianci,  Antonio and Ettorre,  Giovanni Carlo},
  year = {2018},
  month = feb,
  pages = {149–172}
}

@article{thomassin2020ovarian,
  title = {Ovarian-Adnexal Reporting Data System Magnetic Resonance Imaging (O-RADS MRI) Score for Risk Stratification of Sonographically Indeterminate Adnexal Masses},
  volume = {3},
  ISSN = {2574-3805},
  DOI = {https://doi.org/10.1001/jamanetworkopen.2019.19896},
  number = {1},
  journal = {JAMA Network Open},
  publisher = {American Medical Association (AMA)},
  author = {Thomassin-Naggara,  Isabelle and Poncelet,  Edouard and Jalaguier-Coudray,  Aurelie and Guerra,  Adalgisa and Fournier,  Laure S. and Stojanovic,  Sanja and Millet,  Ingrid and Bharwani,  Nishat and Juhan,  Valerie and Cunha,  Teresa M. and Masselli,  Gabriele and Balleyguier,  Corinne and Malhaire,  Caroline and Perrot,  Nicolas F. and Sadowski,  Elizabeth A. and Bazot,  Marc and Taourel,  Patrice and Porcher,  Raphaël and Darai,  Emile and Reinhold,  Caroline and Rockall,  Andrea G.},
  year = {2020},
  month = jan,
  pages = {e1919896}
}

@article{harth2023application,
  title = {Application of the Enzian classification for endometriosis on MRI: prospective evaluation of inter- and intraobserver agreement},
  volume = {10},
  ISSN = {2296-858X},
  DOI = {https://doi.org/10.3389/fmed.2023.1303593},
  journal = {Frontiers in Medicine},
  publisher = {Frontiers Media SA},
  author = {Harth,  Sebastian and Kaya,  Hasan Emin and Zeppernick,  Felix and Meinhold-Heerlein,  Ivo and Keckstein,  J\"{o}rg and Yildiz,  Selcuk Murat and Nurkan,  Emina and Krombach,  Gabriele Anja and Roller,  Fritz Christian},
  year = {2023},
  month = nov 
}

@article{pascoal2022strengths,
  title = {Strengths and limitations of diagnostic tools for endometriosis and relevance in diagnostic test accuracy research},
  volume = {60},
  ISSN = {1469-0705},
  DOI = {https://doi.org/10.1002/uog.24892},
  number = {3},
  journal = {Ultrasound in Obstetrics \&amp; Gynecology},
  publisher = {Wiley},
  author = {Pascoal,  E. and Wessels,  J. M. and Aas‐Eng,  M. K. and Abrao,  M. S. and Condous,  G. and Jurkovic,  D. and Espada,  M. and Exacoustos,  C. and Ferrero,  S. and Guerriero,  S. and Hudelist,  G. and Malzoni,  M. and Reid,  S. and Tang,  S. and Tomassetti,  C. and Singh,  S. S. and Van den Bosch,  T. and Leonardi,  M.},
  year = {2022},
  month = sep,
  pages = {309–327}
}

@inproceedings{baur2019deep,
  title = {Deep Autoencoding Models for Unsupervised Anomaly Segmentation in Brain MR Images},
  ISBN = {9783030117238},
  ISSN = {1611-3349},
  DOI = {https://doi.org/10.1007/978-3-030-11723-8_16},
  booktitle = {Brainlesion: Glioma,  Multiple Sclerosis,  Stroke and Traumatic Brain Injuries},
  publisher = {Springer International Publishing},
  author = {Baur,  Christoph and Wiestler,  Benedikt and Albarqouni,  Shadi and Navab,  Nassir},
  year = {2019},
  pages = {161–169}
}

@article{anchan2024unveiling,
  title = {Unveiling the fibrotic puzzle of endometriosis: An overlooked concern calling for prompt action},
  volume = {13},
  ISSN = {2046-1402},
  DOI = {https://doi.org/10.12688/f1000research.152368.3},
  journal = {F1000Research},
  publisher = {F1000 Research Ltd},
  author = {Anchan,  Megha M and Kalthur,  Guruprasad and Datta,  Ratul and Majumdar,  Kabita and P,  Karthikeyan and Dutta,  Rahul},
  year = {2024},
  month = dec,
  pages = {721}
}

@article{yang2023real,
  title = {Real-Time Automatic Assisted Detection of Uterine Fibroid in Ultrasound Images Using a Deep Learning Detector},
  volume = {49},
  ISSN = {0301-5629},
  DOI = "https://doi.org/10.1016/j.ultrasmedbio.2023.03.013",
  number = {7},
  journal = {Ultrasound in Medicine \&amp; Biology},
  publisher = {Elsevier BV},
  author = {Yang,  Tiantian and Yuan,  Linlin and Li,  Ping and Liu,  Peizhong},
  year = {2023},
  month = jul,
  pages = {1616–1626}
}

@article{chen2020deep,
    title = {Deep learning for the determination of myometrial invasion depth and automatic lesion identification in endometrial cancer MR imaging: a preliminary study in a single institution},
  volume = {30},
  ISSN = {1432-1084},
  DOI = {https://doi.org/10.1007/s00330-020-06870-1},
  number = {9},
  journal = {European Radiology},
  publisher = {Springer Science and Business Media LLC},
  author = {Chen,  Xiaojun and Wang,  Yida and Shen,  Minhua and Yang,  Bingyi and Zhou,  Qing and Yi,  Yinqiao and Liu,  Weifeng and Zhang,  Guofu and Yang,  Guang and Zhang,  He},
  year = {2020},
  month = apr,
  pages = {4985–4994}
}

@inproceedings{shahzad2023automated,
  title = {Automated Uterine Fibroids Detection in Ultrasound Images Using Deep Convolutional Neural Networks},
  volume = {11},
  ISSN = {2227-9032},
  DOI = {https://doi.org/10.3390/healthcare11101493},
  number = {10},
  journal = {Healthcare},
  publisher = {MDPI AG},
  author = {Shahzad,  Ahsan and Mushtaq,  Abid and Sabeeh,  Abdul Quddoos and Ghadi,  Yazeed Yasin and Mushtaq,  Zohaib and Arif,  Saad and ur Rehman,  Muhammad Zia and Qureshi,  Muhammad Farrukh and Jamil,  Faisal},
  year = {2023},
  month = may,
booktitle={Healthcare},
  pages = {1493}
}

@article{huo2023artificial,
  title = {Artificial intelligence-aided method to detect uterine fibroids in ultrasound images: a retrospective study},
  volume = {13},
  ISSN = {2045-2322},
  DOI = {https://doi.org/10.1038/s41598-022-26771-1},
  number = {1},
  journal = {Scientific Reports},
  publisher = {Springer Science and Business Media LLC},
  author = {Huo,  Tongtong and Li,  Lixin and Chen,  Xiting and Wang,  Ziyi and Zhang,  Xiaojun and Liu,  Songxiang and Huang,  Jinfa and Zhang,  Jiayao and Yang,  Qian and Wu,  Wei and Xie,  Yi and Wang,  Honglin and Ye,  Zhewei and Deng,  Kaixian},
  year = {2023},
  month = mar 
}

@article{hodneland2021automated,
  title = {Automated segmentation of endometrial cancer on MR images using deep learning},
  volume = {11},
  ISSN = {2045-2322},
  DOI = {https://doi.org/10.1038/s41598-020-80068-9},
  number = {1},
  journal = {Scientific Reports},
  publisher = {Springer Science and Business Media LLC},
  author = {Hodneland,  Erlend and Dybvik,  Julie A. and Wagner-Larsen,  Kari S. and Šoltészová,  Veronika and Munthe-Kaas,  Antonella Z. and Fasmer,  Kristine E. and Krakstad,  Camilla and Lundervold,  Arvid and Lundervold,  Alexander S. and Salvesen,  Oyvind and Erickson,  Bradley J. and Haldorsen,  Ingfrid},
  year = {2021},
  month = jan 
}

@article{ho2020denoising,
  DOI = {https://doi.org/10.48550/ARXIV.2006.11239},
  author = {Ho,  Jonathan and Jain,  Ajay and Abbeel,  Pieter},
  title = {Denoising Diffusion Probabilistic Models},
  publisher = {arXiv},
  year = {2020},
  copyright = {arXiv.org perpetual,  non-exclusive license},
  journal={Advances in neural information processing systems}
}

@inproceedings{ronneberger2015u,
  title = {U-Net: Convolutional Networks for Biomedical Image Segmentation},
  ISBN = {9783319245744},
  ISSN = {1611-3349},
  DOI = {https://doi.org/10.1007/978-3-319-24574-4_28},
  booktitle = {Medical Image Computing and Computer-Assisted Intervention – MICCAI 2015},
  publisher = {Springer International Publishing},
  author = {Ronneberger,  Olaf and Fischer,  Philipp and Brox,  Thomas},
  year = {2015},
  pages = {234–241}
}

@article{muller2025diffusing,
  DOI = {https://doi.org/10.48550/ARXIV.2508.07903},
  author = {M\"{u}ller,  Johanna P. and Knupfer,  Anika and Bl\"{o}ss,  Pedro and Vittur,  Edoardo Berardi and Kainz,  Bernhard and Hutter,  Jana},
  title = {Diffusing the Blind Spot: Uterine MRI Synthesis with Diffusion Models},
  publisher = {arXiv},
  year = {2025},
  copyright = {Creative Commons Attribution 4.0 International},
  journal={arXiv preprint arXiv:2508.07903},
}

@misc{von-platen-etal-2022-diffusers,
  author = {Patrick von Platen and Suraj Patil and Anton Lozhkov and Pedro Cuenca and Nathan Lambert and Kashif Rasul and Mishig Davaadorj and Dhruv Nair and Sayak Paul and William Berman and Yiyi Xu and Steven Liu and Thomas Wolf},
  title = {Diffusers: State-of-the-art diffusion models},
  year = {2022},
  publisher = {GitHub},
  journal = {GitHub repository},
  url = {https://github.com/huggingface/diffusers}
}

@inproceedings{baugh2023many,
  title = {Many Tasks Make Light Work: Learning to Localise Medical Anomalies from Multiple Synthetic Tasks},
  ISBN = {9783031439070},
  ISSN = {1611-3349},
  DOI = {https://doi.org/10.1007/978-3-031-43907-0_16},
  booktitle = {Medical Image Computing and Computer Assisted Intervention – MICCAI 2023},
  publisher = {Springer Nature Switzerland},
  author = {Baugh,  Matthew and Tan,  Jeremy and M\"{u}ller,  Johanna P. and Dombrowski,  Mischa and Batten,  James and Kainz,  Bernhard},
  year = {2023},
  pages = {162–172}
}

@inproceedings{baugh2024image,
  title = {Image-Conditioned Diffusion Models for Medical Anomaly Detection},
  ISBN = {9783031731587},
  ISSN = {1611-3349},
  DOI = {https://doi.org/10.1007/978-3-031-73158-7_11},
  booktitle = {Uncertainty for Safe Utilization of Machine Learning in Medical Imaging},
  publisher = {Springer Nature Switzerland},
  author = {Baugh,  Matthew and Reynaud,  Hadrien and Marimont,  Sergio Naval and Cechnicka,  Sarah and M\"{u}ller,  Johanna P. and Tarroni,  Giacomo and Kainz,  Bernhard},
  year = {2024},
  month = oct,
  pages = {117–127}
}

@inproceedings{p2023confidence,
  title = {Confidence-Aware and Self-supervised Image Anomaly Localisation},
  ISBN = {9783031443367},
  ISSN = {1611-3349},
  DOI = {https://doi.org/10.1007/978-3-031-44336-7_18},
  booktitle = {Uncertainty for Safe Utilization of Machine Learning in Medical Imaging},
  publisher = {Springer Nature Switzerland},
  author = {P. M\"{u}ller,  Johanna and Baugh,  Matthew and Tan,  Jeremy and Dombrowski,  Mischa and Kainz,  Bernhard},
  year = {2023},
  pages = {177–187}
}

@article{dar2025unconditional,
  DOI = {https://doi.org/10.48550/ARXIV.2402.01054},
  author = {Dar,  Salman Ul Hassan and Seyfarth,  Marvin and Ayx,  Isabelle and Papavassiliu,  Theano and Schoenberg,  Stefan O. and Siepmann,  Robert Malte and Laqua,  Fabian Christopher and Kahmann,  Jannik and Frey,  Norbert and Baeßler,  Bettina and Foersch,  Sebastian and Truhn,  Daniel and Kather,  Jakob Nikolas and Engelhardt,  Sandy},
  title = {Unconditional Latent Diffusion Models Memorize Patient Imaging Data: Implications for Openly Sharing Synthetic Data},
  publisher = {arXiv},
  year = {2024},
  journal={Nature Biomedical Engineering}
}

@article{karimi2020deep,
  title = {Deep learning with noisy labels: Exploring techniques and remedies in medical image analysis},
  volume = {65},
  ISSN = {1361-8415},
  DOI = {https://doi.org/10.1016/j.media.2020.101759},
  journal = {Medical Image Analysis},
  publisher = {Elsevier BV},
  author = {Karimi,  Davood and Dou,  Haoran and Warfield,  Simon K. and Gholipour,  Ali},
  year = {2020},
  month = oct,
  pages = {101759}
}

@article{tripathydeep,
  title={Deep Supervision Attention U-net for segmentation of uterine zones: a multi-center study},
  author={Tripathy, Smiti and Castro, Nyvenn and May, Matthias and Siegler, Lisa and Story, Lisa and Uder, Michael and Hutter, Jana}, 
    DOI = {https://doi.org/10.58530/2025/0105},
note      = {Presented at the ISMRM \& ISMRT Annual Meeting and Exhibition 2025. [Online]. Available: \url{https://archive.ismrm.org/2025/0105.html}}
}

@article{mei2022radimagenet,
  title={RadImageNet: An Open Radiologic Deep Learning Research Dataset for Effective Transfer Learning},
  author={Mei, Xueyan and Liu, Zelong and Robson, Philip M and Marinelli, Brett and Huang, Mingqian and Doshi, Amish and Jacobi, Adam and Cao, Chendi and Link, Katherine E and Yang, Thomas and others},
  journal={Radiology: Artificial Intelligence},
  volume={4},
  number={5},
  pages={e210315},
  year={2022},
  publisher={Radiological Society of North America}
}

@inproceedings{bao2024bmad,
  title={Bmad: Benchmarks for medical anomaly detection},
  author={Bao, Jinan and Sun, Hanshi and Deng, Hanqiu and He, Yinsheng and Zhang, Zhaoxiang and Li, Xingyu},
  booktitle={Proceedings of the IEEE/CVF Conference on Computer Vision and Pattern Recognition},
  pages={4042--4053},
  year={2024}
}

@article{cai2025medianomaly,
  title={Medianomaly: A comparative study of anomaly detection in medical images},
  author={Cai, Yu and Zhang, Weiwen and Chen, Hao and Cheng, Kwang-Ting},
  journal={Medical Image Analysis},
  pages={103500},
  year={2025},
  publisher={Elsevier}
}

\end{document}